\documentclass{article}

\usepackage{PRIMEarxiv}

\usepackage[utf8]{inputenc} 
\usepackage[T1]{fontenc}    
\usepackage{hyperref}       
\usepackage{url}            
\usepackage{booktabs}       
\usepackage{amsfonts}       
\usepackage{nicefrac}       
\usepackage{microtype}      
\usepackage{lipsum}
\usepackage{fancyhdr}       
\usepackage{graphicx}       
\graphicspath{{media/}}     

\title{A Systematic Review on the Detection of Fake News Articles
\thanks{\textit{\underline{Note}}: 
\textbf{ArXiv Pre-Print. Currently submitted to ACM TIST. Awaiting peer-review.}} 
}

\author{
  Nathaniel Hoy, Theodora Koulouri \\
  Brunel University London\\
  Department of Computer Science\\
  United Kingdom\\
  \texttt{\{Nathaniel.Hoy2, Theodora.Koulouri\}@brunel.ac.uk} \\
}

\begin{document}
\maketitle

\begin{abstract}
It has been argued that fake news and the spread of false information pose a threat to societies throughout the world, from influencing the results of elections to hindering the efforts to manage the COVID-19 pandemic. To combat this threat, a number of Natural Language Processing (NLP) approaches have been developed. These leverage a number of datasets, feature extraction/selection techniques and machine learning (ML) algorithms to detect fake news before it spreads. While these methods are well-documented, there is less evidence regarding their efficacy in this domain. By systematically reviewing the literature, this paper aims to delineate the approaches for fake news detection that are most performant, identify limitations with existing approaches, and suggest ways these can be mitigated. The analysis of the results indicates that Ensemble Methods using a combination of news content and socially-based features are currently the most effective. Finally, it is proposed that future research should focus on developing approaches that address generalisability issues (which, in part, arise from limitations with current datasets), explainability and bias. 
\end{abstract}

\section{Introduction}

Fake news can be defined as false stories that aim to intentionally influence its consumers, usually for political gain or for profit through advertising \cite{Allcott0Social}. Historically, fake news was spread largely through word-of-mouth or in print media. With the creation of the world wide web, however, it became possible for everyone to share content globally without regulation or scrutiny \cite{Burkhardt2017Combating}. Crucially, though, discovery of this content was initially limited, as people were forced to either actively search for it, hear about it through word-of-mouth or find it posted on more popular sites. Social media sites, such as Facebook and Twitter, which launched in 2004 and 2006, respectively, removed such barriers in fake news discovery. These sites have made it increasingly easy to distribute fake news to many users through a number of different methods \cite{Posetti0short}. Influential figures may share false news or statements, which are then shared by their large followings; real news posts may be inundated with comments attempting to disprove the truth and faceless pages may share news content that is intentionally misleading. 

Despite evidence that fake news has existed for a number of years online, the term ‘fake news’ only saw a large rise in use following the 2016 US Presidential Election \cite{Sharma2019Combating}.  It has been argued that without fake news targeting the opponents of former President Donald Trump, he would have never won the election \cite{RichardGunther2017Fake}. In more recent times, events such as the riots at the Capitol and the anti-vaxxer movement against the COVID-19 vaccine have suggested that fake news can contribute to the damage of national security and public health \cite{Linden2020Inoculating}. As such, it has been argued that fake news presents a threat to societies and democracies around the world \cite{Lazer2017Combating}.   

As a response, there has been a significant increase in interest within the academic community in how best to address the problem of fake news. For computer scientists, this approach largely focuses on using supervised machine learning (ML) algorithms to identify patterns in fake news and therefore detect them before they are able to spread.   This involves leveraging a labelled dataset, containing both true and false news articles, and using algorithms to determine patterns in these different types of news. These patterns then form a model that aims to predict whether any unseen articles may be true or false. A discussion of these methods can be found in the survey of fake news detection techniques by \cite{Bondielli2019survey}, which highlighted the use of Support Vector Machines (SVMs), Decision Trees and Neural Networks (NNs). These algorithms are also supported by Natural Language Processing (NLP) techniques such as feature extraction and feature selection methods, which create a subset of features and convert text into numerical representations that can be used by these algorithms. These may include techniques such as TF-IDF, Bag of Words and Word2Vec applications of which, are exemplified in \cite{Kaur2020Automating,Najar2019Fake,Gravanis2019Behind} respectively. Bondielli’s paper also presents alternative techniques, such as fact-checking algorithms and diffusion pattern analysis, which could be argued to leverage machine learning methods as part of their respective pipelines.

Despite the existence and application of well-established ML and NLP techniques, fake news detection is an emerging field. As such, research appears not to define fake news in a similar way, which leads to a diversity in how the problem is conceived and addressed. Considering Allcott and Gentzkow's definition that describes fake news as intentionally aiming to mislead its consumers for political or economic reasons, there are other forms of content that have been referred to as fake news, but fall outside this definition. These include satirical news such as the content on the Onion.com \cite{Tandoc2017Digital}, rumours that are not necessarily false \cite{Sharma2019Fake}, and clickbait articles that are being used more by reputable news outlets to drive internet traffic \cite{Chen2015Misleading}.  The heterogeneity in what is considered fake news is further magnified by the different means of distribution of fake news, such as comments on social media, posts and entire websites. As a result, current research is varied in what they define as fake news, with some studies stating that they are looking at ‘fake news’ when in reality they investigate clickbait or satirical news. As the field matures, it is therefore important to clearly delineate  what 'type' of fake news or adjacent concept is being explored. A clear conceptualisation of what fake news is will facilitate the interpretation of the results of each study and will motivate more focused research efforts to address the problem.

Given the variety of definitions of fake news, means of distribution, ML and NLP techniques to address the problem, it is necessary to determine what approaches are being applied and how effective they are for the different types of fake news and means of distribution. This is to inform future research and industrial practice in what techniques are worth further investigation and what techniques are less suitable for this domain. To this end, there is a large number of literature reviews that provide overviews of the different methods of detecting fake news in general \cite{Sharma2019Combating,Elhadad2019Fake,Lahlou2019Automatic,Kaliyar2019Misinformation,Parikh2018Media-Rich,Bondielli2019survey,Hassan2019Survey,Pierri2019False,Guo2020Future,Rana2018Review,Manzoor2019Fake,Klyuev2018Fake}. Much fewer literature reviews also provide insight into the effectiveness of certain methods. However, because these reviews are often not aimed to be systematic or have a different scope, they report the results of a limited number of papers or state the limitations of certain approaches but do not provide an in-depth comparison of the techniques used \cite{Zannettou2019Web,Zhang2020overview,Vishwakarma2020Recent,Hirlekar2020Natural,Zhou2019Fake}. The reviews by \cite{Sharma2019Fake,Mahid2018Fake,Sharma2019Combating} also provide insight to the limitations of particular approaches. The review by \cite{George2020Role} offers a comprehensive comparison of approaches in terms of their effectiveness; however, due to the limited number of studies included in the comparison, the results, while valuable, are not conclusive. \cite{Mahid2018Fake} cites hybrid approaches as being more effective than other approaches and this is somewhat also supported by the review by \cite{Manzoor2019Fake}, which states that "the analysis of fake news content is not sufficient to establish an effective and reliable detection system” and that other aspects of fake news including author and user analysis as well as social context should also be explored. Moreover, literature reviews may often include and compare studies that address different "types" of fake news (rumours, clickbait, social media posts, etc.), as previously mentioned. It could be argued that these forms of fake news have different characteristics, such that different approaches may be more effective. As such, more focused investigations – primary studies and literature reviews – are needed into the suitability of approaches, or combination of approaches, for different types of fake news.

The systematic literature review presented in this paper focuses on intentionally false news articles, omitting social media posts or other means of distribution, as well as other types of fake news. It complements previous comprehensive reviews, by including the most recent research, and by identifying the different methods of fake news detection that have been applied in this specific domain of fake news and providing an in-depth and detailed comparison of their effectiveness.
   
This paper first defines the review protocol, followed by an overview of the literature before reporting on the results pertaining to the research questions of the review. The paper continues with a discussion of the results and concludes by making recommendations for future work in the field.

\newpage
\section{Method}

This review follows the guidelines for systematic literature reviews as described by \cite{Kitchenham2004Procedures}. To better manage the review, the tool ‘Parsifal’\footnote{\href{https://parsif.al/}{https://parsif.al/}} was used. Adhering to Kitchenham’s guidelines, this tool allows researchers to import studies, specify exclusion criteria and write comments regarding reasons for exclusion. It also includes features for carrying out Quality Assessments and Data Extraction. 

\subsection{Research Questions}

Section 1 motivates a focused investigation of the methods for false news detection. As such, this study addresses the following research questions and sub-questions:

\begin{itemize}
    \item \textbf{RQ1.} What methods  are there to detect fake news articles?
    \begin{itemize}
        \item \textbf{RQ1.1} What features are used to detect fake news articles?
        \item \textbf{RQ1.2} What feature selection/extraction approaches are used for training?
        \item \textbf{RQ1.3} What machine learning models are used?
        \item \textbf{RQ1.4} What combination of feature selection/extraction approach is used?
        \item \textbf{RQ1.5} What datasets are used?\\
    \end{itemize}
    \item \textbf{RQ2.} How effective are existing methods to detect fake news articles?
    \begin{itemize}
        \item \textbf{RQ2.1} What groups of features are more effective for detection?
        \item \textbf{RQ2.2} What feature selection/extraction approaches are associated with better performance?
        \item \textbf{RQ2.3} What methods generally perform better overall?
    \end{itemize}
\end{itemize}

RQ1 is intentionally broad to ensure that all methods of detecting fake news are captured. Although other reviews indicate that supervised machine learning is the predominant approach, this review shall attempt to capture other methods that are used, such as unsupervised or semi-supervised approaches (RQ1.3). In addition, it could be argued that the feature extraction/selection approaches, datasets used, and their combination with the chosen ML method, also have a large impact on the resulting performance, and as such, these are also investigated as part of RQ1.1, RQ1.2, RQ1.4 and RQ1.5.  

RQ2 addresses how effective these feature groups, feature extraction/selection approaches and ML methods are at addressing the fake news problem by primarily comparing F-score and accuracy metrics. Due to the number of variables between different papers (including the datasets used and the differing implementations of ML methods NLP techniques), multiple analyses are performed to ensure that more reliable comparisons can be made. This includes comparing the average F-scores and accuracies for each proposed approach, analysing what methods consistently outperform other methods in comparative studies, and investigating performances on the most widely used dataset. 

\subsection{Search Process}

The process to collect relevant studies involved both automated and manual searches. This process is shown diagrammatically in Figure 1.

\subsubsection{Automated Search}

The term ‘fake news’ gained significant popularity during the 2016 US election. As such, articles published in the period between 1st January 2016 and 8th December 2020 were collected. The search process was largely automated by searching databases including IEEEXplore, ACM, ScienceDirect and Scopus.

Using the research questions as a starting point, a process of string refinement was undertaken. Initially, keywords such as ‘fake’, ‘news’ and ‘detect’ along with their synonyms were used to define the search string. However, it became evident that, due to this being a relatively new area of research, a broader search string should be used. This allows the review to be as comprehensive as possible through capturing as much of the literature as possible. Therefore, only synonyms relating to the words ‘fake’ and ‘news’ were used to define the search string, and synonyms relating to ‘detect’ were omitted resulting in the following search string.
\newline

\emph{(Fake OR Misinformation OR False OR Unverified OR Inaccurate OR Rumour/s) AND (News OR Article/s OR Media OR Information)}\\

Due to fake news being an area that spans research areas  outside of Computer Science searches were limited, where possible, to peer-reviewed Computer Science journals and conferences, given the focus of this review on the technical aspects of the fake news problem. Following collection, the majority of duplicates were removed automatically through Parsifal. Duplicates that were not captured by Parsifal were excluded manually upon content review. This stage of data collection is presented in Table 1. 

\begin{table}[h]
\centering
  \caption{Total Papers Collected by Database}
  \label{tab:freq}
  \begin{tabular}{ccl}
    \toprule
    Database&Number of Papers\\
    \midrule
    IEEE Xplore & 236\\
    ACM & 148\\
    Scopus & 577\\
    ScienceDirect & 102\\
  \bottomrule
    \textbf{Total} & 1063\\
    \textbf{After all duplicates removed}&670
\end{tabular}
\end{table}

\subsubsection{Manual Search}

For thoroughness, other relevant literature reviews on this topic were also searched for potential papers to include in this review.  In total, 19 literature reviews were searched with a total of 1137 references. Potential papers were selected based on their title and checked against the automatic search results to ensure that there were no duplicates. Following the selection by title, papers were scanned based on their content and excluded according to the criteria. The papers selected at each stage can be seen in Figure 1. 

\subsection{Study Selection and Evaluation}

Following the collection of papers, a set of exclusion and inclusion criteria was defined in order to filter out papers that were not relevant to the study or did not align with the definitions and research questions of this review. 

\subsubsection{Inclusion Criteria}\hfill
\begin{itemize}
    \item \textbf{IC1.} Computer Science Papers
    \item \textbf{IC2.} Date = 2016 – 2020
    \item \textbf{IC3.} Language = English
    \item \textbf{IC4.} Primary Studies
    \item \textbf{IC5.} Relevant to research questions
\end{itemize}

The inclusion criteria are typical for systematic reviews, whereby studies must be relevant to the research questions and subject area as well as be primary studies. The date range was selected because academic interest in fake news gained significant traction after the 2016 Presidential Election (as discussed in Section 1). Studies were also required to be written in English, such that the authors could understand the content. For studies to be included in the review, they were required to satisfy all the above criteria.

\subsubsection{Exclusion Criteria}\hfill
\begin{itemize}
    \item \textbf{EC1.} Does not focus on news articles.
    \item \textbf{EC2.} Does not address detection of fake news articles.
    \item \textbf{EC3.} Does not present any results for the detection of fake news articles.
    \item \textbf{EC4.} Focuses on a specific event or subject area.
\end{itemize}

The exclusion criteria expand on IC5 by stating what is required for papers to be relevant to the research questions. As discussed in Section 1, the aim of this review is to focus on the different methods to detect intentionally false news articles and their effectiveness (EC1 and EC2). As effectiveness is a key component of this review, results must be presented for comparison (EC3). Finally, this review aims to explore fake news in a topic-agnostic manner, therefore studies should aim to explore detection of fake news in general not in relation to specific topics or events such as COVID-19 (EC4).

\subsubsection{Study Selection}
 
 Initially, papers collected by automated search were excluded based on their title and abstract. This was followed by excluding by their content. Papers collected through a manual search of references from other literature reviews were initially scanned only by their title, followed by a thorough scan on their content. Omitting scanning of the abstract for the manual search was done for practical reasons when searching through literature review references. Figure 1 summarises the result of the study selection process for both the automated search and manual search.
 
 \begin{figure}[h]
  \centering
  \includegraphics[scale=0.5]{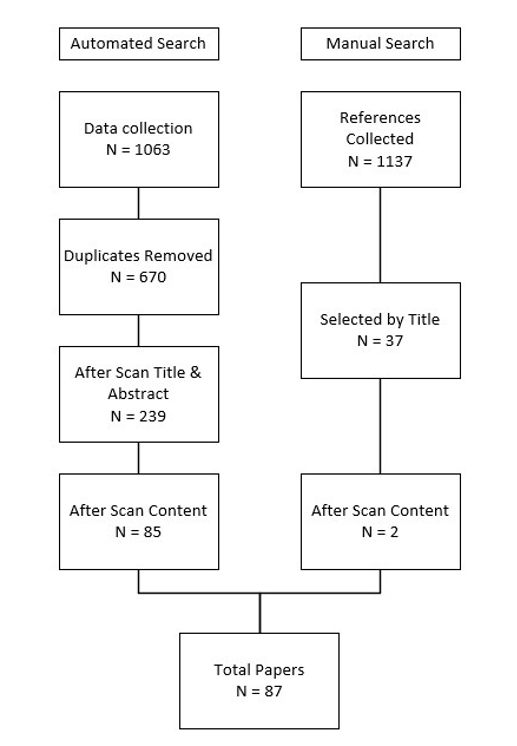}
  \caption{Study Selection Flowchart}
\end{figure}
 
 As can be seen from Figure 1, although numerous papers were extracted from other literature reviews on this topic during the manual search, only two made it through final selection. This provides confidence that the automated search was thorough and captured the majority of relevant papers. 
 
\subsubsection{Quality Assessment}

Following the study selection phase, a Quality Assessment was carried out on the included studies. In systematic reviews, the Quality Assessment can have two purposes:  it is used as a means to either exclude studies or to support data synthesis \cite{Yang2021Quality}. In this study, the Quality Assessment was used to support data synthesis and analysis. This enables the review to more accurately capture the current state of fake news research and identify areas of improvement. The criteria of the Quality Assessment along with an explanation and motivation for each criterion can be found in Table 2. The criteria were formulated as questions, and answers to these questions were restricted to ‘Yes’, ‘Partially’ and 'No', each with a numerical score of 1, 0.5 and 0 given, respectively.

\begin{table}
  \caption{QA Criteria and Justification}
  \label{tab:freq}
  \begin{tabular}{p{0.35\linewidth} | p{0.6\linewidth}}
    \toprule
    Question & Justification\\
    \midrule
    \textbf{QA1.} Does the paper provide an adequate definition or explanation of 'Fake News'? & The term 'fake news' is often used loosely, or as an umbrella term to refer to different forms of intentionally or unintentionally misleading online content (e.g., rumours, satirical content, social media posts, etc.). As such, it is important that research papers define or describe the ‘fake news’ content being addressed.\\\\
    
    \textbf{QA2.} Does the paper disclose/provide access to the dataset used (if applicable)? & The characteristics of the dataset, including size and quality, influences the performance of ML models. To enable comparative evaluation and for the purposes of reproducibility, the dataset used in the study should be identified. \\\\
    
    \textbf{QA3.} Were the contents of the dataset used adequately described (if applicable)? & Similarly, it is important to have an insight into the features used to build the model.  \\\\
    
    \textbf{QA4.} Was the methodology  adequately described? & Here, methodology is defined as the series of steps taken in the NLP pipeline (or if an alternative method has been used, a comprehensive explanation of the steps taken). This is for purposes of reproducibility and evidence of adherence to scientific method. \\\\
    
    \textbf{QA5.} Were one or more performance metrics used with an adequate discussion of results ? & The choice of metric determines how the performance of the model is measured and compared. Disclosure of the metric(s) used and a discussion of the evaluation results provides an insight into the effectiveness of the model, in relation to other approaches, and motivates future research. Here, an ‘adequate discussion’ is defined as an attempt to critically interpret the results and link them to previous work. Papers that fulfil this requirement ‘partially’ are defined as having a discussion section but only enumerating the results presented without any insight to limitations or reliability. Papers that omit a discussion section fail this QA criteria. \\
  \bottomrule
\end{tabular}
\end{table}
\newpage
\subsection{Data Extraction}

\begin{table}
\centering
  \caption{Data Extraction Fields}
  \label{tab:freq}
  \begin{tabular}{ccl}
    \toprule
    Data Extraction Fields\\
    \midrule
    Year of Publication\\
    Authors\\
    Source\\
    Journal/Conference\\
    Title\\
    Dataset Used\\
    Number of Fake Articles\\
    Number of Real Articles\\
    Dataset Size\\
    Users\\
    Features Used\\
    Method Type (Unsupervised, supervised etc.)\\
    Accuracy\\
    F1 Score\\
    Precision\\
    Recall\\
    AUC\\
  \bottomrule
\end{tabular}
\end{table}
The data extraction phase serves to collect data to address the research questions. This was organised by means of a spreadsheet exported from Parsifal where each row contained the selected papers. Appended to this list of papers, a number of attributes were added in relation to the research questions. These are summarised in Table 3. Some fields pertaining to the details of the publication and the authorship were automatically collected through the export process. The remaining fields were filled in manually. The three major groups of data that were manually collected were as follows: the method of detection including the type, algorithm, and features used, addressing RQ1; the performance including metrics such as F1 score, accuracy, precision, recall and Area Under the ROC Curve (AUC), addressing RQ2; finally the dataset used, including its size where possible, in relation to RQ2.

During data extraction, some papers did not directly give the figures for some fields. However, where possible these fields were populated by deriving results from other data collected (for example, F1 score may have been derived using the precision and recall). In cases where studies included results from other papers to be used as a baseline, only the primary results were included in the data extraction to avoid duplication. An exception to this is where a paper repeated another’s method and produced new results through that method. In cases where several results were presented for the same method, with the independent variable not being the method, only the average score was included. 

One prominent issue during the data extraction phase were the variations of the same basic algorithm; an example of this is NuSVM and Linear SVM or variations in the different types of neural networks. These were recorded as they were presented by the selected papers but were also grouped by the algorithms from which they were derived.

Over the course of the data extraction phase, it was observed that the most commonly applied metrics were accuracy and F-score. These will therefore be the main metrics used for comparison between methods. Papers not citing the F-score or accuracy will not be used for direct comparison.

\subsection{Threats to Validity}

As is common in systematic literature reviews, there are a number of threats to validity which may introduce bias into the outcomes of the review. These include publication bias and errors in data collection, study exclusion and data extraction. To mitigate against these threats, the following counter-measures were implemented. In terms of publication bias, whereby studies are more likely to select positive results over negative ones, this is mitigated through the Quality Assessment which attempts to ascertain whether studies discuss their results with limitations. In regards to this review, as the aim is to report the efficacy of different methods in the field, rather than present new results of its own, there is also no motivation from this review to only include studies that report positive results – similar to other SLRs identified by \cite{Kitchenham2010Systematic}. To mitigate against omitting studies based on the search criteria, a broad search string was used as discussed in Section 2.2.1. It could be argued that the date range used could be expanded to studies that were published before 2016; however, as discussed in Section 1, fake news only became popularised from this year onwards. This decision is further justified by the results presented in Section 3.1, which showed a steep increase in publications addressing fake news starting from 2017. Regarding errors in study exclusion and data extraction, where studies may have been incorrectly excluded or the data extraction erroneous, this was mitigated through a review by a secondary researcher. In the initial stages of study selection based on title and abstract, this was carried out by a single author with the sole purpose to only exclude studies that were undoubtedly out of scope (erring on the side of inclusion for any title/abstract that was deemed doubtful). During the selection by content stage, a random sample of papers was taken and reviewed by the secondary author. This approach appears to be the most popular for SLRs as demonstrated by \cite{Carver2013Identifying}, although there is no standard amount of papers to use for this random sample. It was agreed that a significant but manageable number of papers should be undertaken for review by a secondary author, in this case 20\% with an agreement threshold of 90\%. This percentage of papers and agreement between the two researchers was also used for a different random sample in the data extraction stage.

\newpage
\section{Results}
In this section, the results of the study are discussed. Initially, an overview is provided of the included studies. Section 3.2 provides the results of the Quality Assessment. In Section 3.3, results relating to RQ1 are presented about the methods used in the studies – including choice of features, feature selection/extraction method, machine learning model and dataset. Finally, Section 3.4 describes the results relating to RQ2 which focuses on the effectiveness of these approaches.

\subsection{Overview of Included Studies}

This study identified 87 papers from 2016-2020 that were relevant to the research question. 66 of these were from conference proceedings and the remaining 22 were from academic journals. Figure 2 displays where these studies were found, and Figure 3 displays the year in which the studies were published. As can be seen from Figure 3, most selected studies were from later years in the defined range, with no studies being identified in 2016. This supports the decision to keep the selection range between 2016 to 2020. The steep slope in Figure 3 may also indicate that it is a relatively new but also rapidly growing area of interest.

Although the manual search yielded a couple of additional papers, it is also unlikely that a further manual search would have yielded many more studies. The vast majority of the references in the literature reviews searched were either not relevant to the research questions, focused on other areas of fake news (such as social media) or were not experimental studies that would have yielded results in relation to RQ2. The fact that many studies that had been included in these literature reviews were also discovered through the automated search bolsters confidence that this review remained as comprehensive as possible. 

\begin{figure}[h]
\centering
\begin{minipage}{.5\textwidth}
  \centering
  \includegraphics[width=0.84\linewidth]{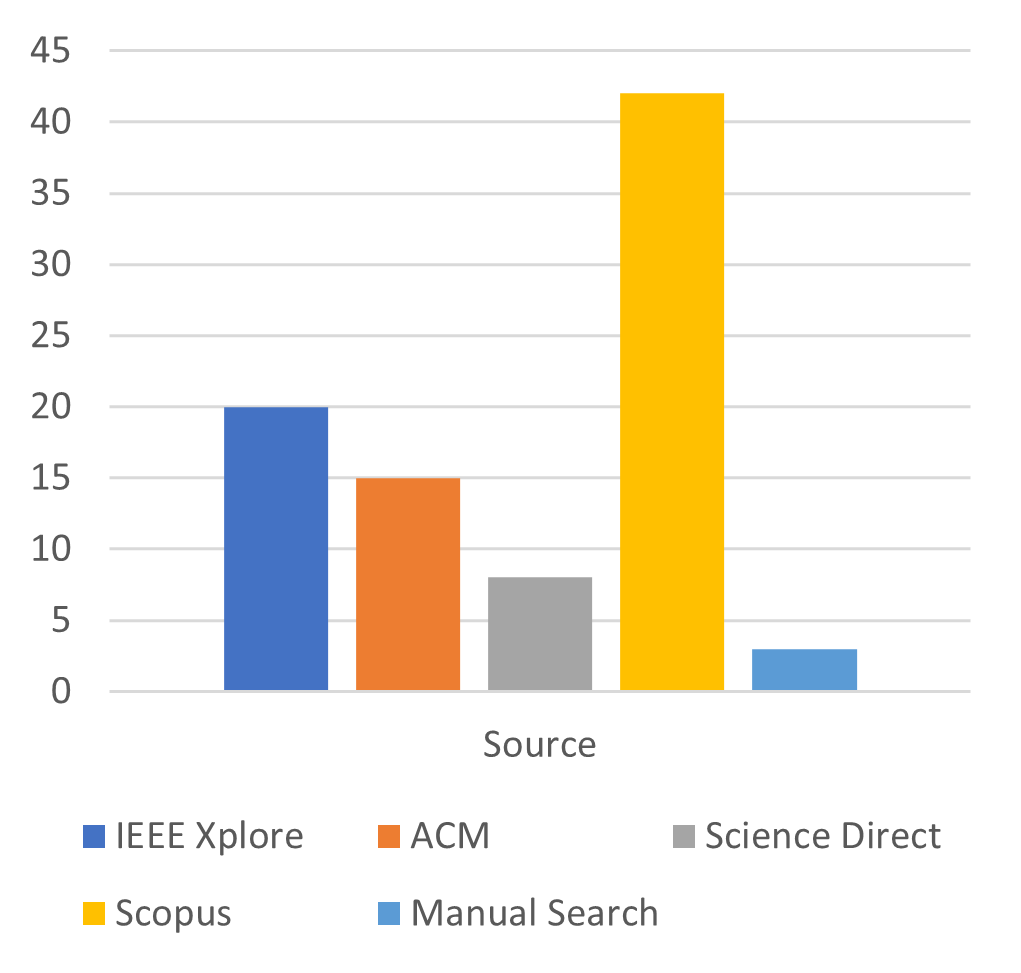}
  \caption{Selected Study Sources}
  \label{fig:test1}
\end{minipage}%
\begin{minipage}{.5\textwidth}
  \centering
  \includegraphics[width=0.8\linewidth]{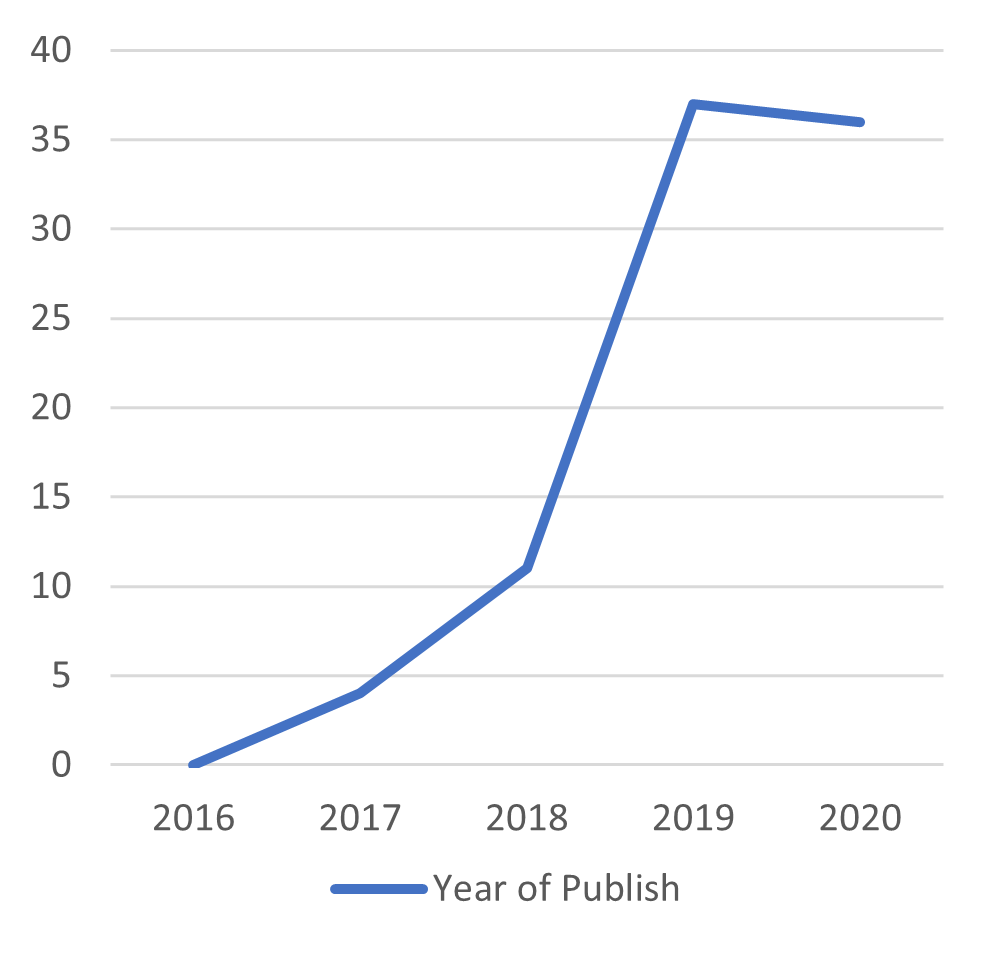}
  \caption{Selected Studies Years of Publish}
  \label{fig:test2}
\end{minipage}
\end{figure}

\subsection{Quality Assessment Results}

A Quality Assessment (QA) of the studies was performed principally to assist in data synthesis as well as to provide insights for future research \cite{Kitchenham2004Procedures}. Five quality assessment questions were derived and can be found in Section 2.3.3.

As can be seen from Figure 4, studies perform relatively well overall with the majority of scores falling above 3. However, looking at individual QA questions associated with lower scores, areas of improvement for future studies can be identified. The results relating to each of the QA questions are presented below. 

\subsubsection{QA1: Definitions of Fake News}

Relating to QA1, it was noted that 41\% of the studies did not provide any definition of ‘Fake News’; rather, these papers discussed the current state of fake news and opportunities in research before moving onto their own approach for solving the problem. As this is a somewhat binary question, only a few studies were marked as ‘partially’ in answering QA1. Studies marked as ‘partially’ answering QA1 were generally studies that alluded to what fake news is, without providing an explicit definition. An example of this can be found in \cite{Reis2019Explainable}, which, in the introduction, explains the impact of fake news which gives the reader some insight to what fake news is but without providing an explicit definition. Lack of clarity in the definitions used in a study may be seen as problematic. As described in Section 1, the study of fake news is an emerging field with no agreed definition of what fake news is. This means that there are deviations in how the fake news problem is being understood and, in turn, being approached and solved.

\subsubsection{QA2: Disclosure and Access to Datasets}

Relating to QA2, 47\% of studies  did at least partially disclose what dataset was used, typically by citing a previous study that has used this dataset while omitting a direct reference to the dataset. A further 37\% disclosed the dataset fully with a direct citation to the dataset used. On the other hand, studies marked as not disclosing the dataset at all were typically studies where a custom dataset was used, which was created by the authors. These studies would largely describe how the dataset was produced, typically through web-scraping and labelling based on where they were scraped from, but would not provide access to the dataset. Disclosing the dataset used could help create performance benchmarks, support transparency and discourage concerns around bias.

\subsubsection{QA3: Dataset Contents}

In relation to QA3, 37\% of studies did not adequately describe the contents of the dataset that was used, particularly in studies that presented models which only trained on textual features. This meant that it was unclear what aspects of a news article were used; for example, whether the headline, author and publication date were used in training. As many of the models are not easily explainable, knowing the contents of the dataset used to train the model could provide some transparency into how a model differentiates between different types of news in the dataset.  

\subsubsection{QA4: Description of Methodologies}

In terms of QA4, the methodology was nearly always adequately described, which is to be expected from studies that rely on experimental work. However, an exception to this was 9 papers that were found not to disclose the feature extraction/selection method that was used. As this is a key step in the NLP pipeline, these papers were marked as partially fulfilling the requirement for this particular QA question. 

\subsubsection{QA5: Disclosure of Metrics and Discussion of Evaluation Results}

The large majority of selected papers provided a clear description of their evaluation metrics and discussion of results, as part of QA5. However, 12 papers were found to satisfy this criterion only partially.  These studies identified the evaluation metrics that they had used and outlined the results, but there was no attempt to interpret the results by linking them to findings of previous studies or state any limitations.

\begin{figure}[h]
\centering
\begin{minipage}{.5\textwidth}
  \centering
  \includegraphics[width=0.8\linewidth]{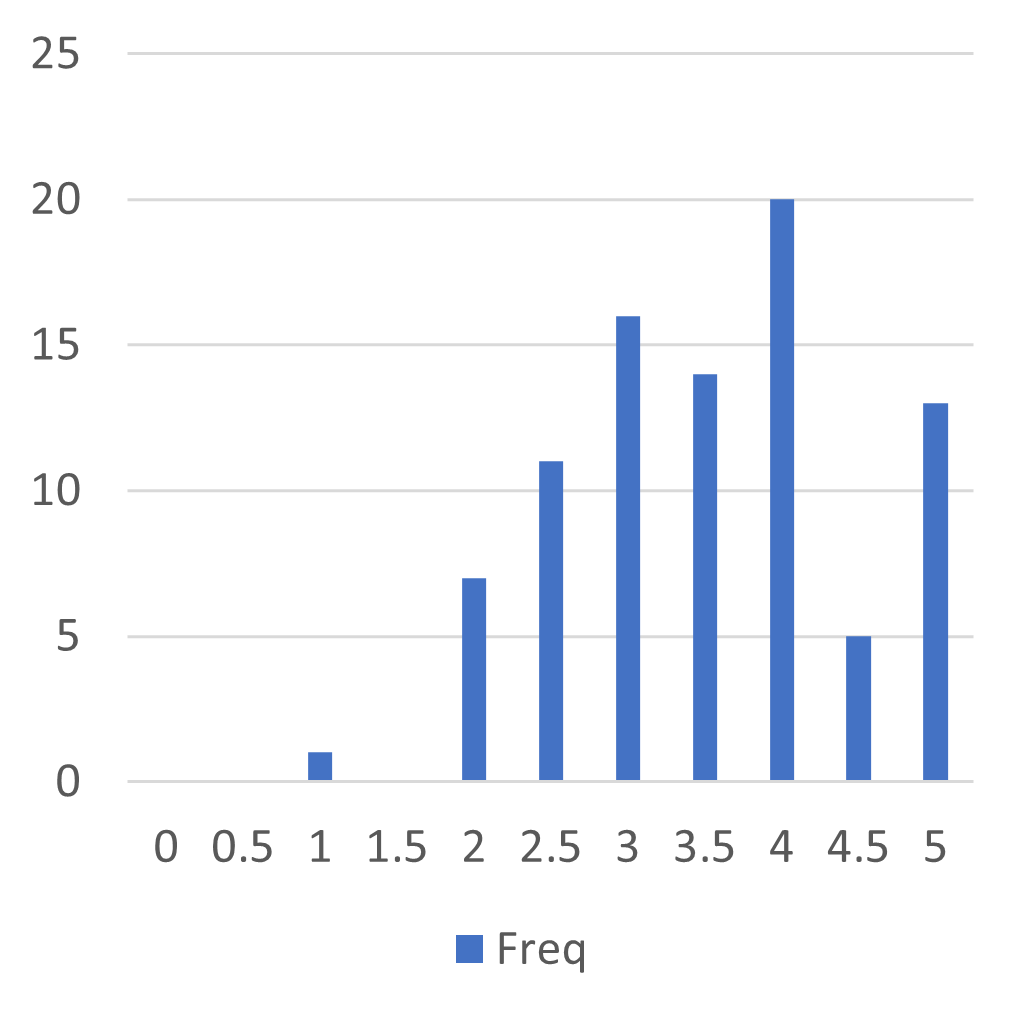}
  \caption{Selected Study Sources}
  \label{fig:test1}
\end{minipage}%
\begin{minipage}{.5\textwidth}
  \centering
  \includegraphics[width=0.81\linewidth]{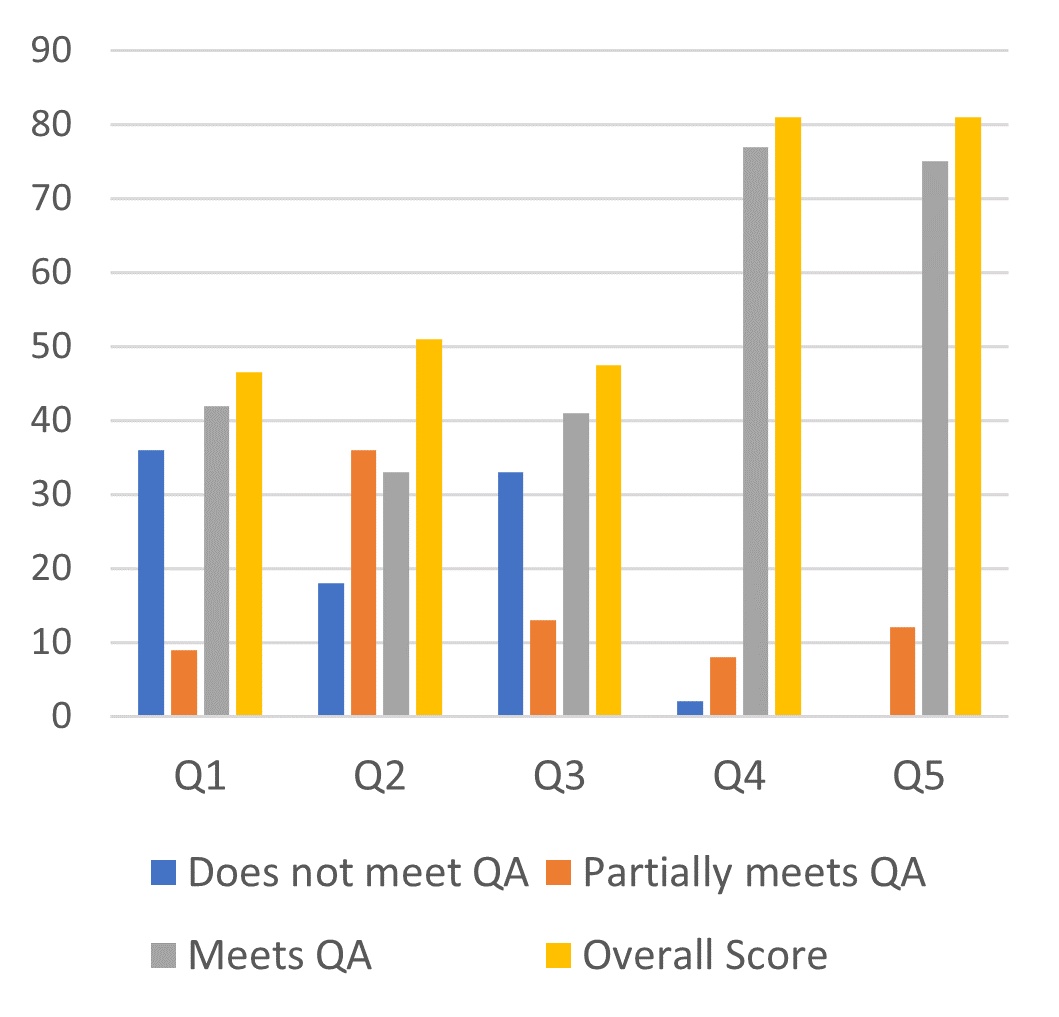}
  \caption{Selected Studies Years of Publish}
  \label{fig:test2}
\end{minipage}
\end{figure}

\subsection{Methods of Fake News Article Detection (RQ1)}

All papers used a supervised approach with the exception of one paper which used a semi-supervised approach. This was despite efforts to attempt to capture other methods as discussed in Section 2. Machine learning approaches for Natural Language Processing problems typically rely on a general pipeline which, among other steps, includes feature selection/extraction, modelling and evaluation; as such, the presentation of the results will follow this pipeline.

\subsubsection{Features (RQ1.1)}

\cite{Xie2020Fake} offers a broad categorisation of three types of approach depending on the features used. These features are as follows: \\
 
\begin{itemize}
        \item	\textbf{News Content-Based:} Features derived from the main body and textual features of the news article which may include the headline, article body, author names and publication date.
        \item	\textbf{Social Context Based:} Features based on data from social media including propagation and comments
        \item	\textbf{Feature Fusion:} Features that are a combination of the first two categories.\\
    \end{itemize}

In the studies selected in this review, content-based features appear to be used significantly more frequently than other types of features (Figure 6). It could be argued that this is because the majority of datasets that are available in this problem domain solely include article content and omit social features. 

\begin{figure}[h]
    \centering
    \includegraphics[scale=0.8]{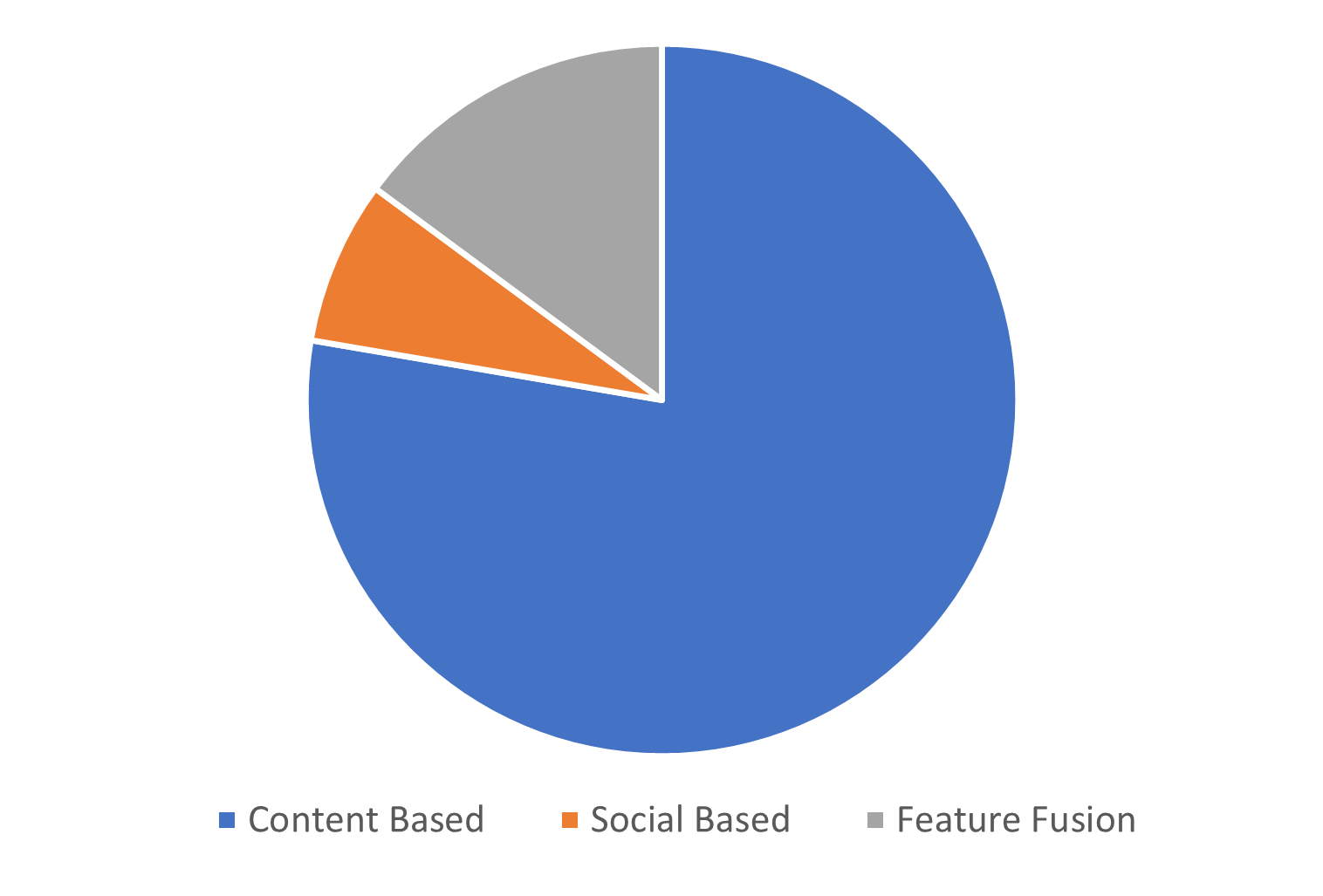}
    \caption{Overview of Feature Groups used by Papers}
    \label{fig:my_label}
\end{figure}

\subsubsection{Feature Selection/Extraction Methods (RQ1.2)}

From a more technical perspective, features in an NLP model can be defined as numerical representations that a machine learning model can use. Prior to forming these numerical representations, it is typical for only a subset of features to be used to train the model. This is usually done in an attempt to improve the performance of the model by selecting only key features and excluding any that may introduce noise into a model. One way to do this is feature selection, which involves including or excluding different parts of the text such as bi-grams, part-of-speech tags, ranges of word-frequencies or named-entities. 
An alternative approach is feature extraction. Unlike feature selection which maintains the original features, feature extraction derives new information from the original features thus transforming them into a new feature subspace. An example of one of these methods is TF-IDF which scores words depending on their frequency in a document as well as their frequency in all documents across the dataset. Words that appear several times in one document, but not across the dataset, are deemed to be highly relevant to the document; words that appear frequently across all documents are deemed less relevant \cite{Rajaraman2011}. This means that very common words in TF-IDF will have values closer to 0, and carry less weight, whereas more document specific words will have values closer to 1.

Figure 7 shows the frequency of use of different feature extraction/selection methods in the selected studies. Although both TF-IDF and Word2Vec are popular extraction methods in general for NLP problems, the latter seems to be a less common method in this particular problem domain. Unlike TF-IDF, which is an extension of the ‘Bag of Words’ approach, Word2Vec uses a neural network to create word associations from text which are then represented in a vector. Words that are semantically related will have similar vector values \cite{Mikolov2013}. When compared to TF-IDF, Word2Vec captures more information about words used in the corpus as opposed to TF-IDF which simply scores a word’s relevancy in their respective documents. However, one weakness of Word2Vec is that it does only capture one sense of a word. For example, Word2Vec may represent the word ‘left’ as the past tense of ‘leave’ or it may represent it as a direction as in the phrase ‘to the left’, but not both, depending on the dictionary of words on which it was trained. Word2Vec is, therefore, context-independent and has a limited vocabulary \cite{Borders2021}. This limits the accuracy of the word associations it represents in its vectors and is unable to derive vectors that are outside its vocabulary. This problem is overcome in the latest state-of-the-art language model developed by Google, BERT \cite{Devlin2018BERT:}. Unlike Word2Vec, BERT is context-dependent and therefore several representations may be made for the same word, depending on the words preceding and succeeding it. It also is able to handle words outside of its vocabulary. Despite these benefits, only one paper in this study used BERT. However, this may be attributed to it being a relatively new approach.

\begin{figure}[h]
    \centering
    \includegraphics[scale=0.7]{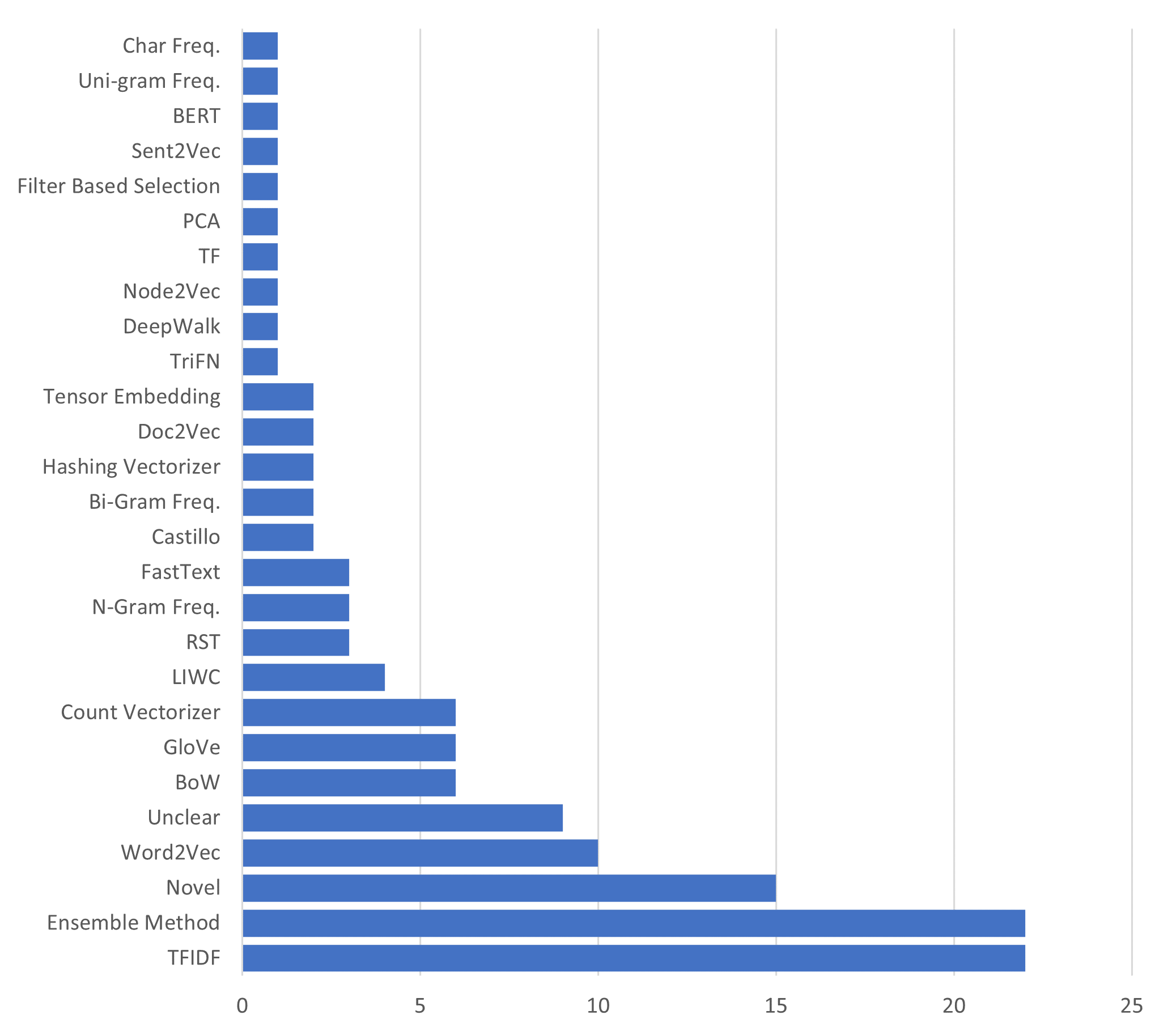}
    \caption{Frequency of Papers using Feature Extraction/Selection Method}
    \label{fig:my_label}
\end{figure}

An equally popular method of feature extraction were Ensemble Methods. Similar to how Ensemble Methods are defined for machine learning models, this is where a number of different approaches are used to extract a combination of features. An example of this can be found in \cite{Mangal2020Fake}, which used a combination of VGGNet to extract features related to an article image, and Word2Vec to extract textual features. To better illustrate these Ensemble Methods, a diagram of the framework used by this paper is provided in Figure 8.

\begin{figure}[h]
    \centering
    \includegraphics[scale=0.5]{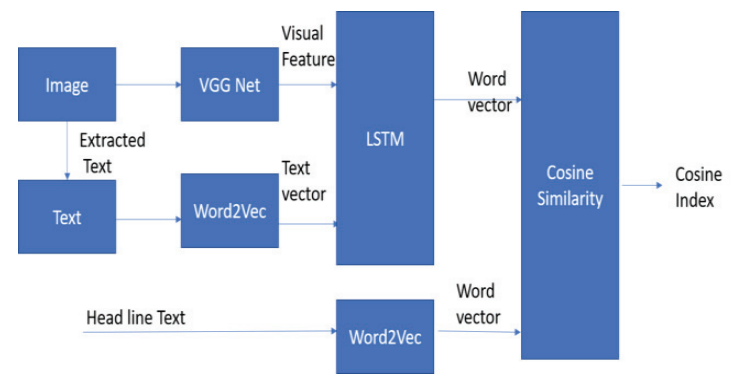}
    \caption{\cite{Mangal2020Fake} Example of Ensemble Feature Extraction Method}
    \label{fig:my_label}
\end{figure}

In 16 cases, a combination of textual and non-textual features were used, and an ensemble of feature extraction methods were applied to combine these features. These features generally included a combination of text, images and social context, as exemplified in \cite{Mangal2020Fake}. The remaining 6 examples of ensemble feature extraction were experiments to see if combining different textual features increased model performance, for example, Word2Vec vectors and TF-IDF scores, as illustrated in \cite{Zhang2018Research}.

It is also worth noting that a number of feature representations were either novel or unclear. 15 papers presented novel feature representation methods, typically using statistical methods that did not relate to well-known approaches such as TD-IDF, Word2Vec or Bag of Words. 9  papers that did not clearly present any feature representation method, as discussed as part of the quality assessment (QA4) in Section 3.2.

\subsubsection{Machine Learning Models (RQ1.3)}

Once a suitable set of features has been derived, they can then be applied to a machine learning model. Figure 9 shows the frequency of different machine learning models in the selected studies. Here, some variations of methods have been grouped to help provide a high-level analysis. For example, different types of gradient boosting such as AdaBoost and XGBoost have been grouped as ‘Gradient Boosting’. Similarly, it is important to note that although Random Forests and Gradient Boosting can be defined as Ensemble Methods in their own right, the term ‘Ensemble Methods’ has been used within this paper to refer to approaches that combine more than one model and do not fall under a well-defined category. This may include combinations of more than one NN trained on different features or a combination of different techniques that use voting to determine the classification. These types of Ensemble Methods also see a fair amount of use within the selected studies. Examples of such approach can be found in \cite{Kaur2020Automating,Ksieniewicz2020Fake}. With this in mind, it could be argued that although NNs are significantly more popular than other approaches listed in Figure 9, all variations of Ensemble Methods, including Random Forests and Gradient Boosting, when combined, are actually more popular.

\begin{figure}[h]
    \centering
    \includegraphics[scale=0.7]{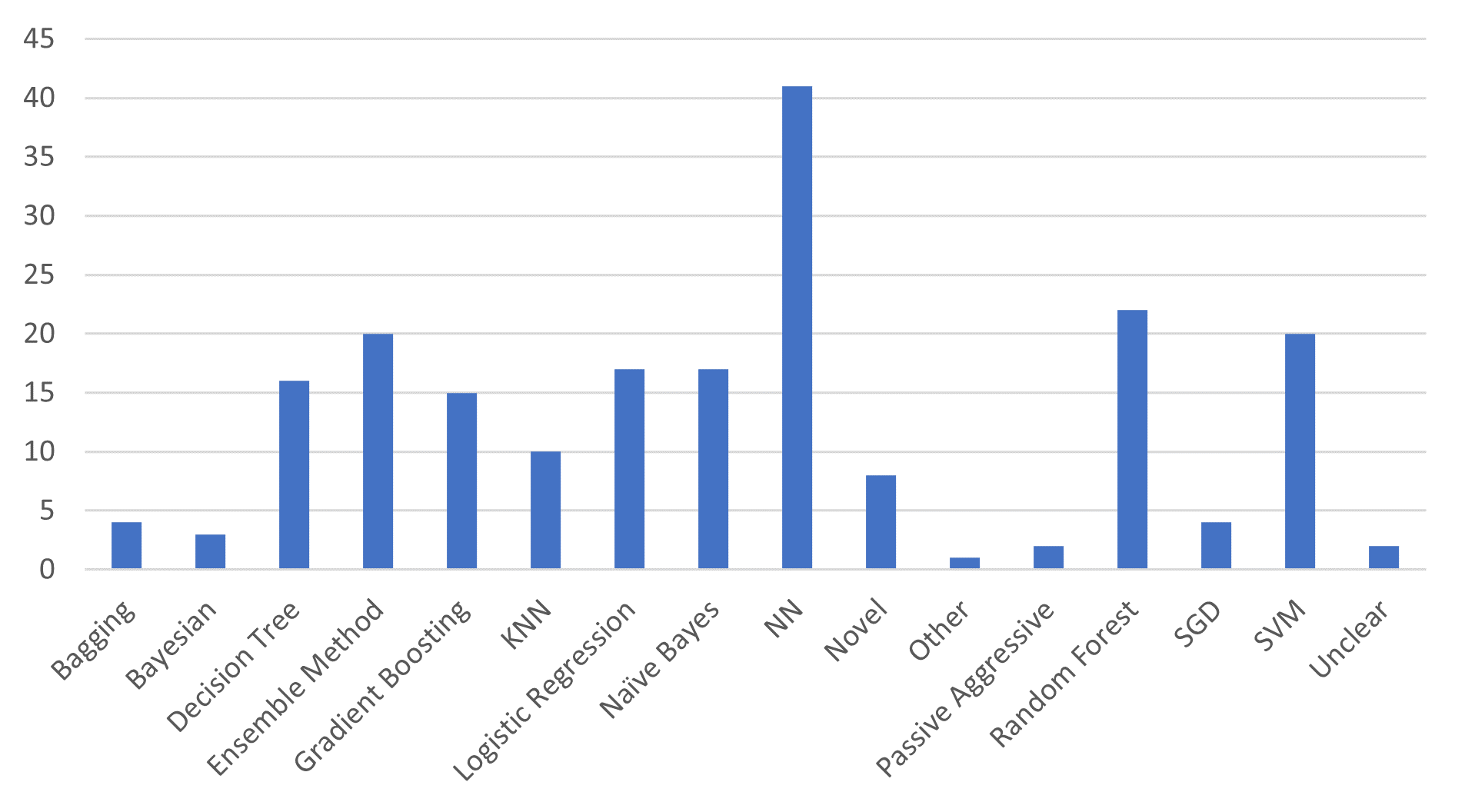}
    \caption{Frequency of Papers Using ML Methods}
    \label{fig:my_label}
\end{figure}

\subsubsection{Feature Extraction and ML Model Combinations (RQ1.4)}


Given that ML models and feature extraction/selection methods both contribute to the efficacy of the resulting model, it is important to investigate what combinations are more commonly used. However, as can be seen from Figure 10, there is no clear combination of technique that is preferred in the literature overall. As in Section 3.2.2, TF-IDF is the most popular feature extraction technique and, as such, it has been attempted more frequently across different machine learning approaches. Due to the amount of overlap between different feature extraction and ML methods, it can be argued that there is currently no agreed combination of techniques. 

\begin{figure}[h]
    \centering
    \includegraphics[scale=0.6]{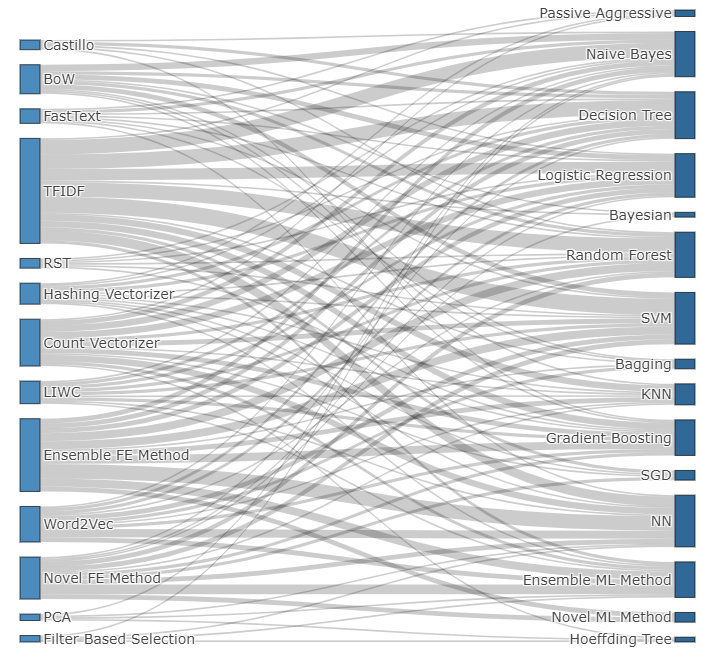}
    \caption{Combinations of ML Methods and Feature Extraction/Selection Methods}
    \label{fig:my_label}
\end{figure}

\subsubsection{Datasets (RQ1.5)}

Another critical parameter is the dataset that was used. Figure 11 shows which datasets were used most by the selected studies and Table 4 provides additional information on the most common publicly available datasets. The key in the Figure 4 lists the datasets in descending order, from most popular datasets to least popular. This analysis indicates that the most popular approach was to use datasets that authors created themselves (labelled ‘Own Creation’ on the chart). This occurred in 21 cases. This is likely due to the relative scarcity of labelled datasets in the Fake News domain. Similarly, authors may have not been satisfied with the quality or size of existing datasets. This may also explain why 19 papers combined two or more datasets together (labelled ‘Combined Dataset’), which was the second most popular approach - often including the 'Getting Real About Fake News' dataset which exclusively includes fake articles, relying on the researcher to gather the real portion themselves.

Of the more established datasets, the FakeNewsNet dataset (and its variations) was the most popular, having been used in 21 papers. Unlike other datasets, this dataset differentiates itself by including several additional features other than the article text such as images, social context and spatiotemporal information. In instances where a paper used features other than article text, this was the dataset that was used. Examples of this include \cite{Cui2019DEFEND:,Chowdhury2020Joint,Xie2020Fake}, all of which leveraged the additional features that FakeNewsNet provides. However despite these additional features, which offers researchers the opportunity to experiment with more novel features, the FakeNewsNet dataset is extremely limited in terms of number of news articles, containing less than 500 articles in total. Table 4 below provides the size of each dataset and the balance between the Fake vs Real news classes.

There is, however, one example of a dataset that is of a size suitable for machine learning. The Fake News Corpus as used by \cite{Paschalides2019Check-It:} contains over ~10 million articles, 3 million of which deemed appropriate for use in this domain by \cite{Paschalides2019Check-It:}. Although this dataset is significant in size, unlike FakeNewsNet the dataset is not manually labelled. Instead, this dataset is coarsely labelled based purely on the domain name that the articles originated from; that is, whether the domain name has been flagged as spreading misinformation. For example, articles that came from ‘Breitbart.com’ would be labelled as ‘fake’ and articles from ‘Telegraph.co.uk’ would be labelled as ‘true’ without evaluating the content of each article independently. This is arguably a rather blunt criterion and introduces noise and bias into the dataset. On closer inspection of the dataset, genuine articles are labelled as fake because of their domain. An example  is the case of the Boston Leader, a website labelled as ‘fake’ by the dataset, which, however contains genuine news articles such as \cite{Minard}. The fact that this dataset was collected by a single author may also present problems, as errors in labelling the domains can occur.

\begin{figure}[h]
    \centering
    \includegraphics[scale=0.7]{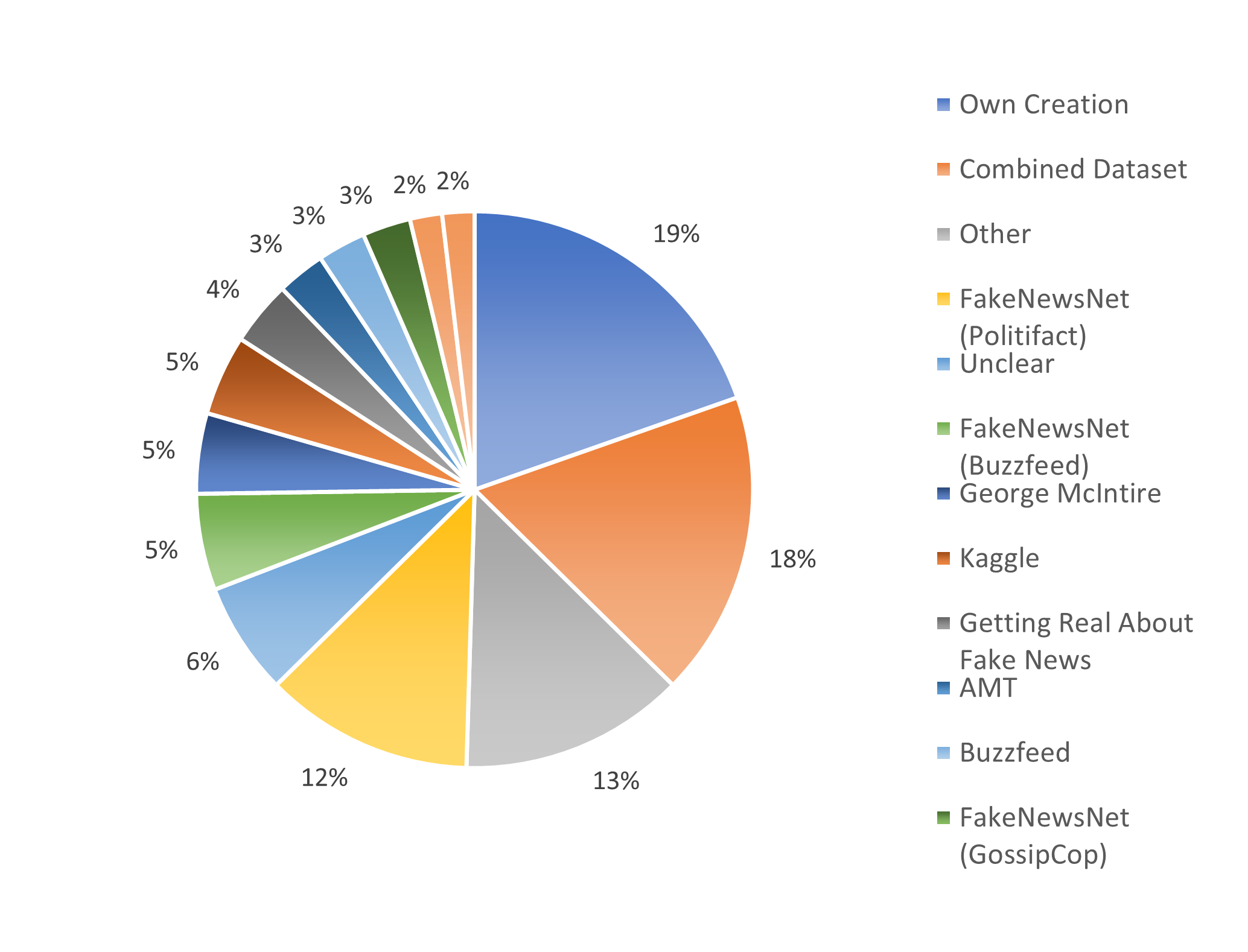}
    \caption{Frequency of Datasets Used by Selected Papers}
    \label{fig:my_label}
\end{figure}

\begin{table}[h]
\centering
  \caption{Metrics of Publicly Available Datasets}
  \label{tab:freq}
  \begin{tabular}{c|c|c|c|c|c}
    \toprule
    Dataset& Number of Times Used &No. Real&No. Fake&Total&Reference\\
    \midrule
    AMT&3&4183&4253&8436&\cite{Janicka2019Cross-domain}\\
    BS Detector&2&12953&15712&28665&\cite{Ahuja2020S-HAN:}\\
    Buzzfeed&3&81&81&162&\cite{Janicka2019Cross-domain}\\
    Fake News Corpus&2&1M&2M&3M&\cite{Paschalides2019Check-It:}\\
    FakeNewsNet (Buzzfeed)&6&91&91&182&\cite{Xie2020Fake}\\
    FakeNewsNet (Politifact)&13&120&120&240&\cite{Xie2020Fake}\\
    George McIntire &5&5279&5279&10558&\cite{Gereme2019Early}\\
    Getting Real About Fake News &3&0&13000&13000&\cite{Katsaros2019Which}\\
    ISOT&3&23481&21417&44898&\cite{Ahmad2020Fake}\\
  \bottomrule
\end{tabular}
\end{table}

\subsection{Effectiveness of Current Methods (RQ2}

During the data extraction phase of this review, it was noted that the two most popular metrics for measuring performance were F-score and accuracy. The F-score is the harmonic mean of the precision and recall of a model. Accuracy is the percentage of correctly predicted cases carried out on an unseen portion of the datasets used. Both range between 0 and 1. Being the most popular metrics, these are the ones that were used to evaluate the effectiveness of different methods. Studies were excluded from the analysis presented in this section if they did not contain these metrics or if these metrics could not be derived from other metrics provided in the studies. Of the 87 studies collected, 14 were excluded for this reason. Of the 73 remaining papers, F-score was deemed a more appropriate measure compared to accuracy, because of the potential of imbalanced class distributions in this problem domain (most news are presumably not fake). This led to 43 being used to summarise the F-score in the remaining 30 papers summarising accuracy respectively in Sections 3.4.1 to 3.4.3. Due to the large amount variance and outliers in the results, the median was used to sort the different methods’ performance as opposed to the mean which is more sensitive to outliers.

\subsubsection{Comparison of Average Performance By Feature Group (RQ2.1)}

The results of the comparison of feature groups in terms of accuracy and F-score are presented in Figure 12. When comparing the feature groups presented in Section 3.2.1, it is interesting to note that the most popular feature group, content-based features, ranks last when comparing against the averages of other feature groups. Arguably, this could be due to the comparatively large number of content-based datasets introducing more variance in the results. Taking into consideration that feature fusion typically contain both content-based and socially-based features, it could be argued that social features have the potential to improve the efficacy of these solutions.   This argument is supported by two papers collected by this review, \cite{Shu2019Beyond,Kaliyar2020DeepFakE:}, which demonstrate how feature fusion and socially-based features outperform content-based features on the same dataset. The comparison between feature groups presented in these two studies is summarised in Figure 13 and reinforces the argument that content-based features are less performant than social and fused features.

\begin{figure}[h]
\centering
\begin{minipage}{.5\textwidth}
  \centering
  \includegraphics[width=1\linewidth]{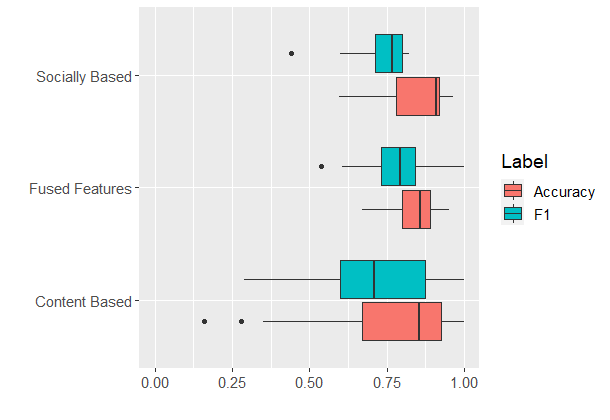}
  \caption{F1 \& Accuracy Scores of Feature Groups}
  \label{fig:test1}
\end{minipage}%
\begin{minipage}{.5\textwidth}
  \centering
  \includegraphics[width=.85\linewidth]{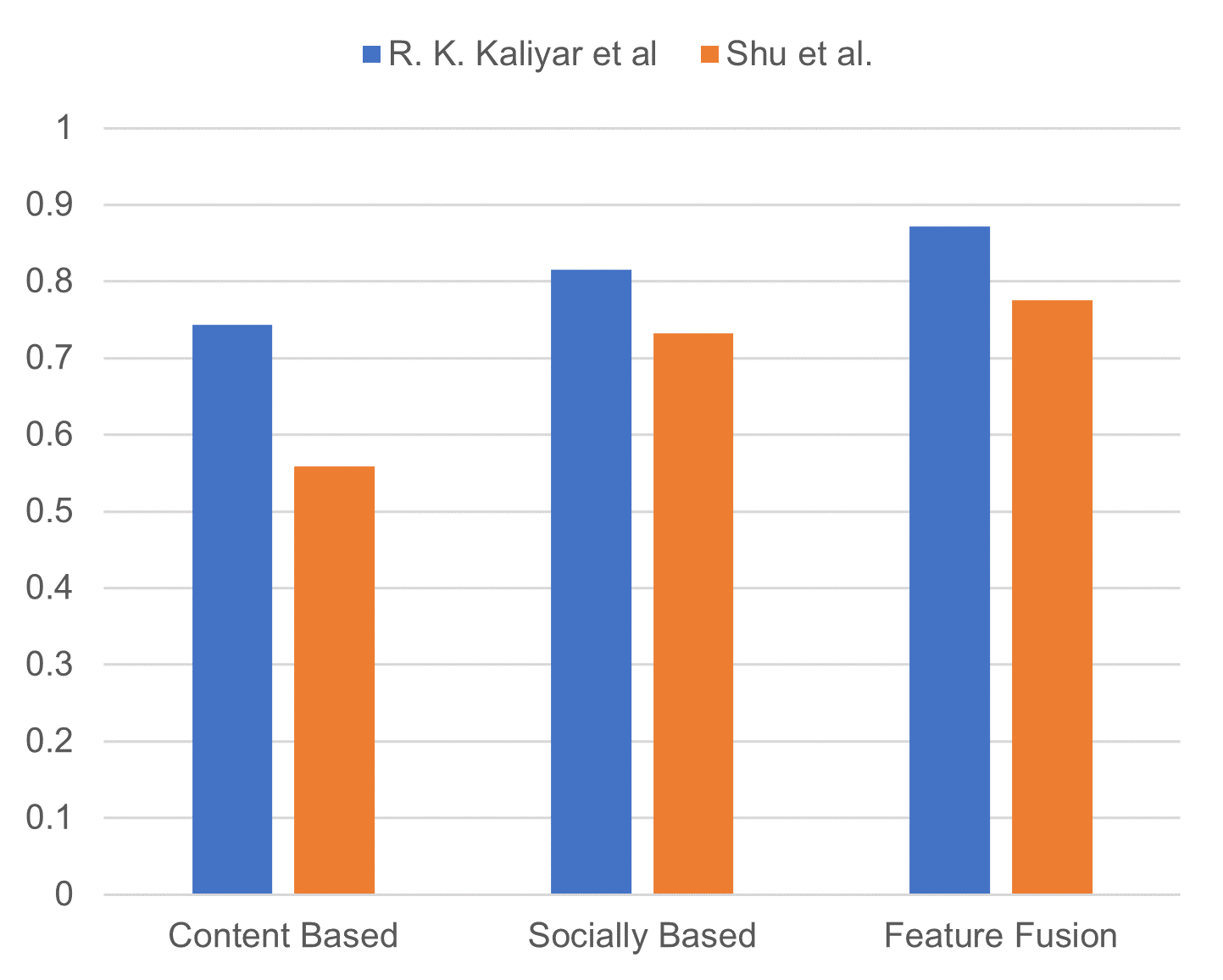}
  \caption{\cite{Kaliyar2019Misinformation,Shu2019Beyond} Feature Group Comparison on FakeNewsNet}
  \label{fig:test2}
\end{minipage}
\end{figure}

\subsubsection{Evaluation of Feature Extraction Approaches (RQ2.2)}

As has been mentioned in this review, the machine learning models employed are not the sole factor that influences the effectiveness of an approach. Along with the dataset, the feature extraction/selection method also plays a key role in determining the efficacy of different approaches. The results are provided in Figures 15 and 16 showing the performance of different feature extraction methods in terms of accuracy and F-score, respectively.

Interestingly, despite TF-IDF and ensemble feature extraction methods being the most popular approach, used in 22 papers of this review, it would appear that they are associated with poorer performances, compared to other feature extraction methods. However, it could be argued that, due to the popularity of the approaches and the amount of variance in the results, this could be the result of the ML methods to which they were applied.

In Section 3.2.3, it was suggested that Word2Vec may be a better approach for this problem, given its ability to capture semantic information about a word, as compared to TF-IDF which simply scores words based on their relevancy to a document in relation to the corpus. It appears here that Word2Vec (along with GloVE, a similar embedding technique) performs well compared to other feature extraction methods. Given that Word2Vec was used in 10 papers in this review, this an encouraging insight that inspires confidence that BERT may produce similar, if not better, results. This is further supported by the fact Tensor embedding techniques also perform very well. However, given these were only used in two papers collected by this review, this needs to be further investigated. Overall, it could be argued that word embeddings is a promising approach to feature extraction for the fake news problem. 

\begin{figure}[h]
\centering
\begin{minipage}{.5\textwidth}
  \centering
  \includegraphics[width=1\linewidth]{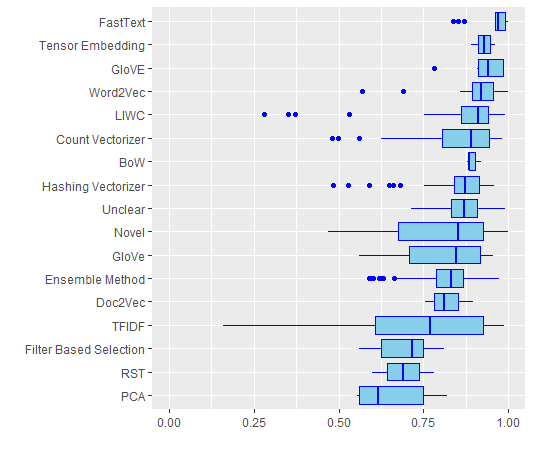}
  \caption{Feature Extraction Performance by Accuracy}
  \label{fig:test1}
\end{minipage}%
\begin{minipage}{.5\textwidth}
  \centering
  \includegraphics[width=1\linewidth]{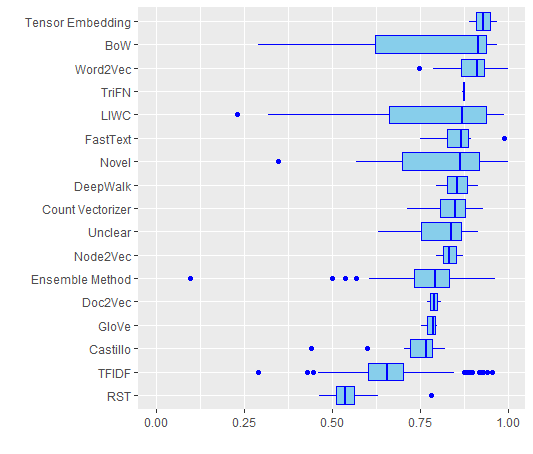}
  \caption{Feature Extraction Performance by F-Score}
  \label{fig:test2}
\end{minipage}
\end{figure}

\newpage
\subsubsection{Comparison of Average Performance by ML Method (RQ2.3)}

Figure 17 and 18 show the average performance of the ML methods reported in the selected studies measured as accuracy and F-score, respectively. When comparing different ML algorithms, it could be argued that most algorithms perform relatively well with most having a median above 0.75 across both metrics/groups of papers. Across the two metrics,  a fair amount of similarity in terms of the ranking of different methods is observed.  Neural networks appear to perform well across both metrics/groups of papers as well as Gradient Boosting algorithms. Logistic Regression also sees similarity in performance across the two metrics/groups along with Random Forests. However these two methods appear to be less performant compared to NNs and Gradient Boosting. It is also worth noting that KNN ranks worst among both metrics/groups and it could be argued that this demonstrates that this is not an effective approach to the problem. This could be due to KNN’s tendency to assume that training data is equally distributed across classes, which may not always be true when random training subsets or inbalanced datasets are used \cite{Tan2006}.

Despite much similarity in the ranking of methods between the two metrics/groups of papers, there are a few notable exceptions to this. When comparing the F-score group to the Accuracy group, we see Ensemble Methods drop from the highest ranked method by F-ccore to the second lowest ranked in accuracy. On closer inspection, this is likely due to the paper ‘Fake News Detection from Data Streams’ \cite{Ksieniewicz2020Fake} which reported accuracy scores for numerous combinations of algorithms, many of which did not perform as well as other Ensemble approaches reported by other papers, thus bringing the median down in the Accuracy group.  This is a limitation of the analysis presented in this Section, and the results should be interpreted accordingly.  Due to the number of variables including datasets, approaches, algorithms and metrics, it is difficult to make conclusive comparisons on the different approaches, which is a common problem in the ML field, however, as reported by \cite{Demsar2006Statistical}. Other approaches that also saw a difference in ranking greater than two places across the two metrics included SVMs and Naïve Bayes. Although this may also be because of the limitation of this analysis as outlined above, the drop in ranking between these two algorithms in the F-score group may also be attributed to this metric penalising false positives and negatives more than accuracy does.

\begin{figure}[h]
\centering
\begin{minipage}{.5\textwidth}
  \centering
  \includegraphics[width=1\linewidth]{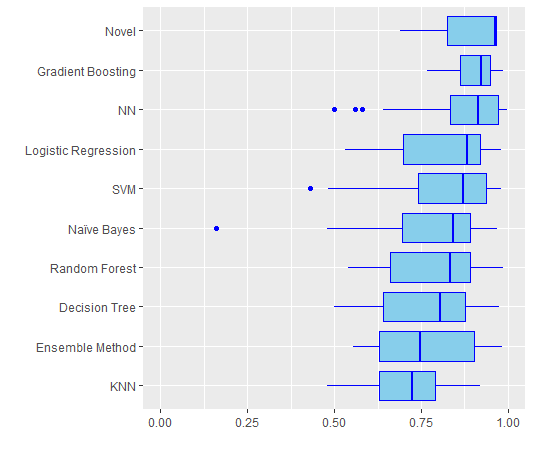}
  \caption{ML Model Performance by Accuracy}
  \label{fig:test1}
\end{minipage}%
\begin{minipage}{.5\textwidth}
  \centering
  \includegraphics[width=1\linewidth]{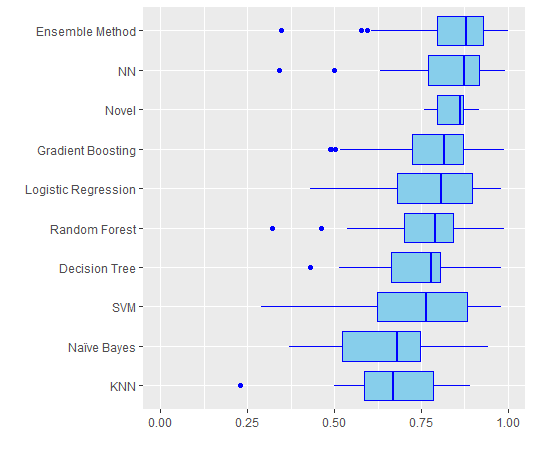}
  \caption{ML Model Performance by F-Score}
  \label{fig:test2}
\end{minipage}
\end{figure}

\newpage
\subsubsection{Comparison by Consistent Outperformance  (RQ2.3)}

In order to address the limitation of the analysis for comparing ML methods presented in the previous section, additional analysis was performed following the approach presented in \cite{Uddin0Comparing}, which used the number of times that an algorithm outperformed other algorithm(s) within the same paper. In addition to this, the analysis reported in this section calculates the number of times a particular algorithm outperforms other algorithms within the same paper as well as on the same dataset. In particular, papers that report comparative results between two or more algorithms were used – this totalled 37 papers. That is, papers that did not compare algorithms (such as the ones that presented a single novel method) were not included in this analysis. In addition, only methods that were reported in more than 5 papers were included. The results are summarised in Table 5.

In this analysis, we see that Gradient Boosting and Ensemble Methods rank joint-first outperforming other algorithms 50\% of the time. This is to be expected, as the paper in the Accuracy group that solely dealt with Ensemble Methods and likely skewed the performance of this method across F-score and accuracy in the previous section was excluded from this analysis, given that it was not a comparative study. Assuming this paper to be an outlier, it could be argued that Ensemble Methods do appear to be a particularly effective approach. In contrast, we see that KNN is the least performant once again and could be regarded as a potentially poor approach to the fake news problem. Naïve Bayes also sees a poor performance in this analysis, with only one occurrence of it outperforming another algorithm. It could be argued that this singular instance is less significant, as the study in which it outperformed other algorithms only contained one other algorithm – Random Forest. With this in mind and its relatively poor performance in the F-score comparison, it could be said that Naïve Bayes may also be less effective for this particular problem. This is supported by the fact that SVMs performed relatively well in this comparison, despite suffering a similar drop in rank to Naïve Bayes when comparing F-score and accuracy.\\

\begin{table}[h]
\centering
  \caption{Frequency of ML Method Outperformance in Comparative Studies}
  \label{tab:freq}
  \begin{tabular}{c|c|c|c}
    \toprule
    Method&Number of Datasets&Superior Performance Freq.&Percentage\\
    \midrule
    Gradient Boosting&22&11&50\%\\
    Ensemble Method&12&6&50\%\\
    Neural Networks&30&10&33\%\\
    SVM&27&7&26\%\\
    Random Forest&29&6&21\%\\
    Decision Tree&23&3&13\%\\
    Logistic Regression&24&2&8\%\\
    Naive Bayes&20&1&5\%\\
    KNN&13&0&0\%\\
  \bottomrule
\end{tabular}
\end{table}

\newpage
\subsubsection{FakeNewsNet Analysis (RQ2.3)}

As identified in Section 3.3.1, the reliability of comparisons between ML methods is limited by the variability of approaches used in studies. The analysis presented in Section 3.3.2 mitigated some of this variability by aggregating the results of studies that compare different algorithms on the same dataset. To further reduce variability, additional analysis was performed to include only studies that used the most popular dataset, the Politifact portion of the FakeNewsNet dataset. Although this does not remove all the variables across studies and limits the sample size to 13 papers, it does offer the opportunity to investigate the effectiveness of the methods using a ‘baseline dataset’. Unlike other datasets, FakeNewsNet offers additional features outside of the text including images and social data. 

As can be seen from Figure 19, similar to the comparisons presented in Sections 3.3.1 and 3.3.2, Ensemble Methods perform best overall with Naïve Bayes and KNN being the least performant. It is also worth noting that the other, well-defined Ensemble Methods such as Gradient Boosting and Random Forests also see good performance, similar to that seen in the previous comparisons. Neural networks however, appear to be less performant on this dataset.
Of the 10 papers that utilised the additional, social features, three of them utilised Ensemble Methods, with the remaining seven utilising Logistic Regression, Gradient Boosting, Neural Networks and KNN. Broadly, it would appear that there are two distinct approaches when using a combination of textual and non-textual data features:

\begin{itemize}
    \item Using more than one classifier (Ensemble Methods) and interpreting the combined results, typically through an additional classifier or through a voting mechanism. 
    \item Using an ensemble of feature extraction approaches and applying the result to a single classifier. 
\end{itemize}

Of these two approaches, it is argued that the first is the most performant given how Ensemble Methods rank top of this comparison. When looking at the previous comparisons, it could be argued that Ensemble Methods produce the best results even when training only on textual features. Papers that used Ensemble Methods on solely textual features typically extracted additional secondary features from the text. Examples of this include semantic analysis and syntax analysis as used by \cite{Zhang2019Detecting}, or voting classifiers to check for agreement across several approaches as used by   \cite{Kaur2020Automating}. Given this performance, it could be argued that article text alone is not enough for the fake news problem and additional features should either be sought from the dataset or derived from textual data. This also supports the suggestion in Section 3.2.3 that word embedding methods such as Word2Vec and BERT should be utilised more widely given the extra contextual information that they provide.\\

\begin{figure}[h]
    \centering
    \includegraphics[scale=0.6]{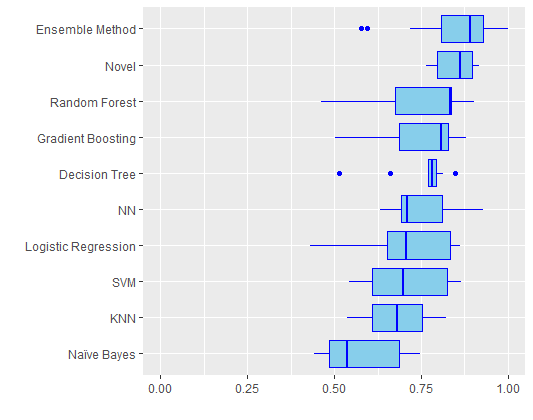}
    \caption{Performance of Methods Applied to FakeNewsNet by F-Score}
    \label{fig:my_label}
\end{figure}

\newpage
\section{Discussion}
Fake news detection is a relatively new field, as can be seen from the substantial increase in publications, as presented in Section 2.1, particularly over the last two years. This growth in interest has led to a variety of approaches to solving the fake news problem and results appear to be promising. Supporting these approaches are also a number of datasets, many of which are created by the authors. The diversity of approaches and datasets indicates that this research is still in its infancy with no well-established set of approaches and baseline datasets, and while there are solutions that produce positive results (particularly from NNs, Gradient Boosting and Ensemble Methods), further focused experimentation and validation efforts are needed before these solutions can be scaled. 

 One of the fundamental issues within the problem area is the lack of a generally accepted and clear definition of what is classed as ‘fake news’. As discussed in Section 1,  clickbait, rumours, satire or verifiably false articles have invariably been referred to as fake news in literature \cite{Bondielli2019survey}. This was also observed during the study selection process, in which several studies define their focus to be fake news but, on closer inspection, they were found to deal exclusively with clickbait articles. Due to these varying definitions, it has been argued that implementation of these models will lead to AI bias concerns and arguments that it will also undermine democracy and infringe free speech \cite{Rainie2017Future}. Given these issues, it could be argued that it may be impossible to create a model that satisfies everyone’s definition of fake news across different topics. However, that does not exclude such models from being applicable to certain situations and remain useful, for example, for social media companies which are under increasing pressure to police their platforms. 

If models are to be applied in real-world scenarios, they must be accurate, robust and generalisable. A key component to achieving this is the size and quality of available datasets. As mentioned in Section 3.2.4, mainstream datasets are relatively small for such a machine learning problem. This could be a reason as to why many authors combined datasets, in order to boost the number of articles for model training and to protect against issues such as overfitting, poor generalisability as well as bias (as justified in \cite{Telang2019Anempirical,Uppal2020Fake}). Existing annotation approaches, manual vs. automated annotation, have limitations which result in a trade-off; that is, datasets are: either well-labelled but smaller and expensive in terms of human labour, by using independent fact-checkers, or poorly labelled but larger using generic labelling algorithms, such as labelling based on the source domain name of the articles. To underscore the amount of effort required, Kaggle’s Getting Real About Fake News dataset at only 13,000 articles is also not manually labelled, instead using a tool called ‘BS Detector’ to label news by its domain. Ultimately, this approach is problematic as BS Detector is a web-browser extension that attempts to detect unreliable articles by comparing hyperlinks on articles to a list of unreliable sources \cite{Oshikawa2020Survey}. As there is very little evidence to BS Detector’s accuracy in labelling data, it could be argued that this approach leads to a domino effect, whereby an unproven model labels data poorly, which in turn is used to train classifiers. It is clear therefore, that in order for a high-quality dataset of significant size to be developed, it would require the continuous resource investment in manual labelling, or the development of better algorithms, such as reliable unsupervised approaches as explored in \cite{Gangireddy2020Unsupervised}. 

Despite apparent issues in current datasets, algorithms appear to perform relatively well, as can been seen in Section 3, with typical F1 and accuracy scores around 0.8. In particular, ensemble methods (including Gradient Boosting and Random Forest) appear to perform best overall, particularly, when trained on a variety of features. The effectiveness of utilising more than just news-content based features is evident in Section 3.3.1, which shows that models that utilise socially-based features together with content-based features typically perform best overall. It is worth noting that although TF-IDF is the most popular means of extracting textual features, word embedding algorithms such as Word2Vec and FastText are associated with better results, likely due to their ability to preserve the context of words unlike TF-IDF. 

However, despite the promising performance of certain combinations of machine learning model, dataset and feature extraction method there is one clear issue that prevents these models from being applied in the real-world. That is, the poor generalisability of the models as demonstrated by \cite{Janicka2019Cross-domain,Gautam2020SGG:}. These studies show that models that are trained on one fake news dataset and then tested on an alternative fake news dataset suffer a drop in accuracy that mirrors random classification. It is possible that this is a result of the limitations of current datasets, as previously discussed, or it could be possible that generalisability is a weakness of content-based approaches as used in these two studies. Models that utilise both social and content-based features may prove to be more resilient to this generalisability issue. Further to this, there are two other generalisability considerations. One is the performance of a model over time as demonstrated by \cite{Horne2019Robust}. Although the drop in performance experienced in this study is not significant, it is still worth consideration given the rate at which news, and therefore language, changes - particularly over longer periods of time than explored in Horne's paper. Additionally, although every attempt has been made by this review to select papers that attempt to detect news across different topics, many of the datasets used solely cover one area of news (such as politics in the FakeNewsNet Politifact dataset). Although these datasets cover a number of different events and topics within this area of news, exploration of how well these models can be applied across various topics (such as healthcare) would be insightful to future research, and assist in determining whether the inclusion of topic-modelling would be beneficial.

Similar to other applications of machine learning in other domains (such as in healthcare), it could be argued that applying these algorithms on news is unethical. This is because machine learning algorithms do not differentiate based on what truly makes a news article fake, but, instead, they derive patterns based on the contents of the fake and real portions of the datasets. It is important therefore, that future approaches are able to classify fake news in an explainable manner, such as through fact-checking algorithms built on knowledge bases \cite{Huynh2019,Sharma2019Combating}. The advantage of such an approach would be that they would potentially be more robust to the kinds of attacks highlighted by \cite{Horne2019Robust} as well as be more generalisable and perform better over time. This is because, unlike current approaches, fact-checking approaches classify based on statements that can be verified to be true or false. However, these approaches are still in their infancy and it could be argued that building and maintaining such knowledge bases would require a large amount of effort – similar to the effort required to build the datasets currently used for this problem. Furthermore, ensuring that the knowledge bases are populated with accurate information is an additional consideration. An alternative approach may also be to use ‘human-in-the-loop' (HITL) approaches as proposed by \cite{Demartini2020Human-in-the-loop} whereby current approaches are used to flag potentially false content before being moderated by a human expert. An advantage of this approach is that it utilises currently available approaches collected by this review and through moderation with human-experts, will allow more data to be labelled which is likely to create more effective models.

\section{Future Work and Conclusions}
In this paper, a systematic literature review was performed in order to determine what current approaches there are to detect fake news articles and how effective they are. First, a combination of automated and manual searches was carried out using a broad search string to capture as many papers related to the research questions as possible, and a set of exclusion and inclusion criteria were applied. Second, a quality assessment was carried out on the 87 selected papers that highlighted some common issues. Third, data extraction was performed to collect the different types of ML models, datasets, feature extraction methods and performance metrics. Finally, a narrative analysis and discussion of the data collected was presented which identified a number of issues relating to current research in the area. 

The review demonstrates a growing interest in this area, particularly over the last two years. It also shows current popular approaches, datasets and feature extraction methods. Some of the more popular approaches include NNs, SVMs, Gradient Boosting, Decision Trees and Ensemble Methods, typically supported by ensemble, word embedding or TF-IDF approaches to feature extraction. Many datasets have been applied to this problem, typically containing content-based features but also social-context based features such as in FakeNewsNet. Overall, study quality is relatively good however there are some common issues, such as lack of clear definition of fake news as well as poor disclosure of textual features used. 

Of the 87 papers collected, the vast majority present relatively promising present promising average accuracy of 0.8. Of the most promising approaches, methods that utilise ensemble learning with a combination of features appear to perform the best with word embeddings being indicated as the best feature extraction method for text, coupled with additional features such as social context found in the FakeNewsNet dataset. 
 
This review also revealed areas of future research.  Chief among these areas is the issue of generalisability.  This has three strands: (i) generalisability from one dataset to the other; (ii) generalisability over time; and (iii) generalisability from one domain to the other.  Preliminary results from \cite{Janicka2019Cross-domain} show that ensemble methods may also be the most favourable to address the second strand of the generalisability issue however, social-context based features are yet to be explored in terms of this strand of generalisability. 
Finally, the ethical implications of current methods should also be considered in future research. This may be addressed through alternative methods such as fact-checking algorithms which form a chain of evidence, or through HITL approaches that ensure that detection results are moderated by a human expert.

\newpage
\bibliographystyle{unsrt}  
\bibliography{references}

\begin{thebibliography}{10}

\bibitem{Allcott0Social}
Hunt Allcott and Matthew Gentzkow.
\newblock Social media and fake news in the 2016 election.
\newblock 0.
\newblock Defines fake news.

\bibitem{Burkhardt2017Combating}
Joanna~M Burkhardt.
\newblock Combating fake news in the digital age history of fake news.
\newblock Technical Report~8, 11 2017.

\bibitem{Posetti0short}
Julie Posetti and Alice Matthews.
\newblock A short guide to the history of 'fake news' and disinformation a
  learning module for journalists and journalism educators.
\newblock 0.

\bibitem{Sharma2019Combating}
Karishma Sharma, Feng Qian, He~Jiang, Natali Ruchansky, Ming Zhang, and Yan
  Liu.
\newblock Combating fake news: A survey on identification and mitigation
  techniques.
\newblock {\em ACM Trans. Intell. Syst. Technol.}, 10(3), 4 2019.

\bibitem{RichardGunther2017Fake}
By~Richard~Gunther, Paul~A Beck, Erik~C Nisbet, Philip~N Howard, Gillian
  Bolsover, Bence Kollanyi, Samantha Bradshaw, Lisa-Maria Neudert, and Junk
  News.
\newblock Fake news did have a significant impact on the vote in the 2016
  election.
\newblock {\em COMPROP Data Memo}, 2017.

\bibitem{Linden2020Inoculating}
Sander van~der Linden, Jon Roozenbeek, and Josh Compton.
\newblock Inoculating against fake news about covid-19.
\newblock {\em Frontiers in Psychology}, 11, 10 2020.

\bibitem{Lazer2017Combating}
David Lazer, Matthew Baum, Nir Grinberg, Lisa Friedland, Kenneth Joseph, Will
  Hobbs, Carolina Mattsson, Yochai Benkler, Adam Berinsky, Helen Boaden,
  Katherine Brown, Gordon Pennycook, Lori Robertson, David Rothschild, Steven
  Sloman, Cass Sunstein, Emily Thorson, and Duncan Watts.
\newblock Combating fake news: An agenda for research and action.
\newblock {\em Filippo Menczer}, 2017.

\bibitem{Bondielli2019survey}
Alessandro Bondielli and Francesco Marcelloni.
\newblock A survey on fake news and rumour detection techniques.
\newblock {\em Information Sciences}, 497:38--55, 2019.

\bibitem{Kaur2020Automating}
Sawinder Kaur, Parteek Kumar, and Ponnurangam Kumaraguru.
\newblock Automating fake news detection system using multi-level voting model.
\newblock {\em Soft Computing}, 24(12):9049--9069, 6 2020.
\newblock cited By 3.

\bibitem{Najar2019Fake}
F~Najar, N~Zamzami, and N~Bouguila.
\newblock Fake news detection using bayesian inference.
\newblock pages 389--394, 2019.

\bibitem{Gravanis2019Behind}
G~Gravanis, A~Vakali, K~Diamantaras, and P~Karadais.
\newblock Behind the cues: A benchmarking study for fake news detection.
\newblock {\em Expert Systems with Applications}, 128:201--213, 2019.
\newblock cited By 14.

\bibitem{Tandoc2017Digital}
Edson~C Tandoc, Wei Lim, and Richard Ling.
\newblock Digital journalism defining "fake news" a typology of scholarly
  definitions.
\newblock 2017.

\bibitem{Sharma2019Fake}
S~Sharma and D~K Sharma.
\newblock Fake news detection: A long way to go.
\newblock pages 816--821, 2019.

\bibitem{Chen2015Misleading}
Yimin Chen, Niall~J Conroy, and Victoria~L Rubin.
\newblock Misleading online content: Recognizing clickbait as "false news".
\newblock 2015.

\bibitem{Elhadad2019Fake}
M~K Elhadad, K~Fun Li, and F~Gebali.
\newblock Fake news detection on social media: A systematic survey.
\newblock pages 1--8, 2019.

\bibitem{Lahlou2019Automatic}
Yasmine Lahlou, Sanaa El~Fkihi, and Rdouan Faizi.
\newblock Automatic detection of fake news on online platforms: A survey.
\newblock {\em ICSSD 2019 - International Conference on Smart Systems and Data
  Science}, pages 2019--2022, 2019.

\bibitem{Kaliyar2019Misinformation}
R~K Kaliyar and N~Singh.
\newblock Misinformation detection on online social media-a survey.
\newblock pages 1--6, 2019.

\bibitem{Parikh2018Media-Rich}
S~B Parikh and P~K Atrey.
\newblock Media-rich fake news detection: A survey.
\newblock pages 436--441, 2018.

\bibitem{Hassan2019Survey}
E~A Hassan and F~Meziane.
\newblock A survey on automatic fake news identification techniques for online
  and socially produced data.
\newblock pages 1--6, 2019.

\bibitem{Pierri2019False}
F~Pierri and S~Ceri.
\newblock False news on social media: A data-driven survey.
\newblock {\em SIGMOD Record}, 48(2):18--32, 2019.
\newblock cited By 7.

\bibitem{Guo2020Future}
Bin Guo, Yasan Ding, Lina Yao, Yunji Liang, and Zhiwen Yu.
\newblock The future of false information detection on social media: New
  perspectives and trends.
\newblock {\em ACM Computing Surveys}, 53(4), 2020.

\bibitem{Rana2018Review}
D~P Rana, I~Agarwal, and A~More.
\newblock A review of techniques to combat the peril of fake news.
\newblock pages 1--7, 2018.

\bibitem{Manzoor2019Fake}
S~I Manzoor, J~Singla, and Nikita.
\newblock Fake news detection using machine learning approaches: A systematic
  review.
\newblock pages 230--234, 2019.

\bibitem{Klyuev2018Fake}
V~Klyuev.
\newblock Fake news filtering: Semantic approaches.
\newblock pages 9--15, 2018.

\bibitem{Zannettou2019Web}
Savvas Zannettou, Michael Sirivianos, Jeremy Blackburn, and Nicolas Kourtellis.
\newblock The web of false information: Rumors, fake news, hoaxes, clickbait,
  and various other shenanigans.
\newblock {\em J. Data and Information Quality}, 11(3), 5 2019.

\bibitem{Zhang2020overview}
X~Zhang and A~A Ghorbani.
\newblock An overview of online fake news: Characterization, detection, and
  discussion.
\newblock {\em Information Processing and Management}, 57(2), 2020.
\newblock cited By 33.

\bibitem{Vishwakarma2020Recent}
D~K Vishwakarma and C~Jain.
\newblock Recent state-of-the-art of fake news detection: A review.
\newblock pages 1--6, 2020.

\bibitem{Hirlekar2020Natural}
Vaishali~Vaibhav Hirlekar and Arun Kumar.
\newblock Natural language processing based online fake news detection
  challenges – a detailed review.
\newblock (Icces):748--754, 2020.

\bibitem{Zhou2019Fake}
Xinyi Zhou, Reza Zafarani, Kai Shu, and Huan Liu.
\newblock Fake news: Fun-damental theories, detection strategies and
  challenges.
\newblock 19, 2019.

\bibitem{Mahid2018Fake}
Z~I Mahid, S~Manickam, and S~Karuppayah.
\newblock Fake news on social media: Brief review on detection techniques.
\newblock pages 1--5, 2018.

\bibitem{George2020Role}
J~George, S~M Skariah, and T~A Xavier.
\newblock Role of contextual features in fake news detection: A review.
\newblock pages 1--6, 2020.

\bibitem{Kitchenham2004Procedures}
Barbara Kitchenham.
\newblock Procedures for performing systematic reviews.
\newblock {\em Keele, UK, Keele University}, 33(2004):1--26, 2004.

\bibitem{Yang2021Quality}
Lanxin Yang, He~Zhang, Haifeng Shen, Xin Huang, Xin Zhou, Guoping Rong, and
  Dong Shao.
\newblock Quality assessment in systematic literature reviews: A software
  engineering perspective.
\newblock {\em Information and Software Technology}, 130:106397, 2 2021.

\bibitem{Kitchenham2010Systematic}
Barbara Kitchenham, Rialette Pretorius, David Budgen, O~Pearl~Brereton, Mark
  Turner, Mahmood Niazi, and Stephen Linkman.
\newblock Systematic literature reviews in software engineering-a tertiary
  study.
\newblock 2010.

\bibitem{Carver2013Identifying}
Jeffrey~C. Carver, Edgar Hassler, Elis Hernandes, and Nicholas~A. Kraft.
\newblock Identifying barriers to the systematic literature review process.
\newblock {\em International Symposium on Empirical Software Engineering and
  Measurement}, pages 203--213, 2013.

\bibitem{Reis2019Explainable}
Julio C~S Reis, André Correia, Fabr{\textbackslash}'{\textbackslash}icio
  Murai, Adriano Veloso, and Fabr{\textbackslash}'{\textbackslash}icio
  Benevenuto.
\newblock Explainable machine learning for fake news detection.
\newblock WebSci '19, page 17–26, New York, NY, USA, 2019. Association for
  Computing Machinery.

\bibitem{Xie2020Fake}
Yi~Xie, Xixuan Huang, Xiaoxuan Xie, and Shengyi Jiang.
\newblock A fake news detection framework using social user graph.
\newblock BDE 2020, page 55–61, New York, NY, USA, 2020. Association for
  Computing Machinery.

\bibitem{Rajaraman2011}
Anand Rajaraman and Jeffrey~David Ullman.
\newblock {Data Mining}.
\newblock {\em Mining of Massive Datasets}, pages 1--17, 2011.

\bibitem{Mikolov2013}
Tomas Mikolov, Kai Chen, Greg Corrado, and Jeffrey Dean.
\newblock {Efficient Estimation of Word Representations in Vector Space}.
\newblock {\em 1st International Conference on Learning Representations, ICLR
  2013 - Workshop Track Proceedings}, jan 2013.

\bibitem{Borders2021}
Tammie~L Borders and Svitlana Volkova.
\newblock {An Introduction to Word Embeddings and Language Models}.
\newblock apr 2021.

\bibitem{Devlin2018BERT:}
Jacob Devlin, Ming-Wei Chang, Kenton Lee, and Kristina Toutanova.
\newblock Bert: Pre-training of deep bidirectional transformers for language
  understanding.
\newblock {\em NAACL HLT 2019 - 2019 Conference of the North American Chapter
  of the Association for Computational Linguistics: Human Language Technologies
  - Proceedings of the Conference}, 1:4171--4186, 10 2018.

\bibitem{Mangal2020Fake}
Deepak Mangal and Dilip~Kumar Sharma.
\newblock Fake news detection with integration of embedded text cues and image
  features.
\newblock pages 68--72. IEEE, 6 2020.

\bibitem{Zhang2018Research}
S~Zhang, Y~Wang, and C~Tan.
\newblock Research on text classification for identifying fake news.
\newblock pages 178--181, 2018.

\bibitem{Ksieniewicz2020Fake}
Pawel Ksieniewicz, Pawel Zyblewski, Michal Choras, Rafal Kozik, Agata Gielczyk,
  and Michal Wozniak.
\newblock Fake news detection from data streams.
\newblock pages 1--8. IEEE, 7 2020.

\bibitem{Cui2019DEFEND:}
Limeng Cui, Kai Shu, Suhang Wang, Dongwon Lee, and Huan Liu.
\newblock Defend: A system for explainable fake news detection.
\newblock CIKM '19, page 2961–2964, New York, NY, USA, 2019. Association for
  Computing Machinery.

\bibitem{Chowdhury2020Joint}
Rajdipa Chowdhury, Sriram Srinivasan, and Lise Getoor.
\newblock Joint estimation of user and publisher credibility for fake news
  detection.
\newblock CIKM '20, page 1993–1996, New York, NY, USA, 2020. Association for
  Computing Machinery.

\bibitem{Paschalides2019Check-It:}
D~Paschalides, C~Christodoulou, R~Andreou, G~Pallis, M~D Dikaiakos,
  A~Kornilakis, and E~Markatos.
\newblock Check-it: A plugin for detecting and reducing the spread of fake news
  and misinformation on the web.
\newblock pages 298--302, 2019.

\bibitem{Minard}
Minard.
\newblock {A Round-up of 2019 Boston Marathon-Related Charitable Activities -
  Boston Leader}, 2019.

\bibitem{Janicka2019Cross-domain}
M~Janicka, M~Pszona, and A~Wawer.
\newblock Cross-domain failures of fake news detection.
\newblock {\em Computacion y Sistemas}, 23(3):1089--1097, 2019.
\newblock cited By 0.

\bibitem{Ahuja2020S-HAN:}
N~Ahuja and S~Kumar.
\newblock S-han: Hierarchical attention networks with stacked gated recurrent
  unit for fake news detection.
\newblock pages 873--877, 2020.

\bibitem{Gereme2019Early}
Fantahun~Bogale Gereme and William Zhu.
\newblock Early detection of fake news "before it flies high".
\newblock ICBDT2019, page 142–148, New York, NY, USA, 2019. Association for
  Computing Machinery.

\bibitem{Katsaros2019Which}
D~Katsaros, G~Stavropoulos, and D~Papakostas.
\newblock Which machine learning paradigm for fake news detection?
\newblock pages 383--387, 2019.

\bibitem{Ahmad2020Fake}
I~Ahmad, M~Yousaf, S~Yousaf, and M~O Ahmad.
\newblock Fake news detection using machine learning ensemble methods.
\newblock {\em Complexity}, 2020, 2020.
\newblock cited By 0.

\bibitem{Shu2019Beyond}
Kai Shu, Suhang Wang, and Huan Liu.
\newblock Beyond news contents: The role of social context for fake news
  detection.
\newblock WSDM '19, page 312–320, New York, NY, USA, 2019. Association for
  Computing Machinery.

\bibitem{Kaliyar2020DeepFakE:}
R~K Kaliyar, A~Goswami, and P~Narang.
\newblock Deepfake: improving fake news detection using tensor
  decomposition-based deep neural network.
\newblock {\em Journal of Supercomputing}, 2020.
\newblock cited By 0.

\bibitem{Tan2006}
Songbo Tan.
\newblock {An effective refinement strategy for KNN text classifier}.
\newblock {\em Expert Systems with Applications}, 30(2):290--298, feb 2006.

\bibitem{Demsar2006Statistical}
Janez Demšar.
\newblock Statistical comparisons of classifiers over multiple data sets.
\newblock Technical report, 2006.

\bibitem{Uddin0Comparing}
Shahadat Uddin, Arif Khan, Ekramul Hossain, and Mohammad~Ali Moni.
\newblock Comparing different supervised machine learning algorithms for
  disease prediction.
\newblock 0.

\bibitem{Zhang2019Detecting}
C~Zhang, A~Gupta, C~Kauten, V~A Deokar, and X~Qin.
\newblock Detecting fake news for reducing misinformation risks using analytics
  approaches.
\newblock {\em European Journal of Operational Research}, 279(3):1036--1052,
  2019.
\newblock cited By 18.

\bibitem{Rainie2017Future}
Lee Rainie, Janna Anderson, and Jonathan Albright.
\newblock The future of free speech, trolls, anonymity and fake news online,
  2017.
\newblock [Online; accessed 2021-06-07].

\bibitem{Telang2019Anempirical}
H~Telang, S~More, Y~Modi, and L~Kurup.
\newblock Anempirical analysis of classification models for detection of fake
  news articles.
\newblock pages 1--7, 2019.

\bibitem{Uppal2020Fake}
A~Uppal, V~Sachdeva, and S~Sharma.
\newblock Fake news detection using discourse segment structure analysis.
\newblock pages 751--756, 2020.

\bibitem{Oshikawa2020Survey}
Ray Oshikawa, Jing Qian, and William~Yang Wang.
\newblock A survey on natural language processing for fake news detection.
\newblock 2020.

\bibitem{Gangireddy2020Unsupervised}
Siva Charan~Reddy Gangireddy, Deepak P, Cheng Long, and Tanmoy Chakraborty.
\newblock Unsupervised fake news detection: A graph-based approach.
\newblock HT '20, page 75–83, New York, NY, USA, 2020. Association for
  Computing Machinery.

\bibitem{Gautam2020SGG:}
Akansha Gautam and Koteswar~Rao Jerripothula.
\newblock Sgg: Spinbot, grammarly and glove based fake news detection.
\newblock pages 174--182. IEEE, 9 2020.

\bibitem{Horne2019Robust}
Benjamin~D Horne, Jeppe N{\textbackslash}orregaard, and Sibel Adali.
\newblock Robust fake news detection over time and attack.
\newblock {\em ACM Trans. Intell. Syst. Technol.}, 11(1), 12 2019.

\bibitem{Huynh2019}
Viet-Phi Huynh and Paolo Papotti.
\newblock {A Benchmark for Fact Checking Algorithms Built on Knowledge Bases}.
\newblock In {\em Proceedings of the 28th ACM International Conference on
  Information and Knowledge Management}, volume~10, pages 689--698, New York,
  NY, USA, nov 2019. ACM.

\bibitem{Demartini2020Human-in-the-loop}
Gianluca Demartini, Stefano Mizzaro, and Damiano Spina.
\newblock Human-in-the-loop artificial intelligence for fighting online
  misinformation: Challenges and opportunities.
\newblock (September):65--74, 2020.

\end{thebibliography}

\end{document}